\documentclass[10pt,twocolumn,letterpaper]{article}

\usepackage{cvpr}
\usepackage{times}
\usepackage{epsfig}
\usepackage{graphicx}
\usepackage{amsmath}
\usepackage{amssymb}
\usepackage{textcomp,gensymb}
\usepackage{dblfloatfix}
\usepackage{tabulary}
\usepackage[numbers]{natbib}
\graphicspath{{figures/}}
\usepackage{tabulary}
\usepackage{multirow}
\usepackage[labelfont=bf,textfont=it]{caption}
\usepackage{flushend}

\usepackage[pagebackref=true,breaklinks=true,colorlinks,bookmarks=false]{hyperref}

\cvprfinalcopy % *** Uncomment this line for the final submission

\def\cvprPaperID{8} % *** Enter the CVPR Paper ID here

\newcommand{\comment}[1]{}

\renewcommand{\dblfloatsep}{0.05in}

\renewcommand{\dbltextfloatsep}{0.05in}

\ifcvprfinal\fi
\begin{document}

%%%%%%%%% TITLE
\title{Learning Localized Geometric Features Using 3D-CNN: An Application to Manufacturability Analysis of Drilled Holes }

\author{Aditya Balu\\
\and Sambit Ghadai\\
\and Kin Gwn Lore\\
\and Gavin Young\\
\and \space\\
\and Adarsh Krishnamurthy\\
\and Soumik Sarkar\\
\and Department of Mechanical Engineering, Iowa State University\\
Ames, Iowa, 50011, USA\\
{\tt\small {(baditya, sambitg, kglore, gyoungis, adarsh, soumiks)}@iastate.edu}
}

\comment
{
1. Introduction:
How this is related to computer vision is what we need to add...
We already information on the reason for why manufacturability is important and how ML solves it.
2. Representation of geometry, and how differential geometry can help in representing the geometry
3. 3D Convolutional neural network
4. DFM Rules and Test set explanation
5. Grad-CAM
6. Results and Discussion
7 Conclusion and future works
}

\maketitle
%\thispagestyle{empty}

%%%%%%%%% ABSTRACT
\begin{abstract}
3D convolutional neural networks (3D-CNN) have been used for object recognition based on the voxelized shape of an object. In this paper, we present a 3D-CNN based method to learn distinct local geometric features of interest within an object. In this context, the voxelized representation may not be sufficient to capture the distinguishing information about such local features. To enable efficient learning, we augment the voxel data with surface normals of the object boundary. We then train a 3D-CNN with this augmented data and identify the local features critical for decision-making using 3D gradient-weighted class activation maps. An application of this feature identification framework is to recognize \emph{difficult-to-manufacture} drilled hole features in a complex CAD geometry.  The framework can be extended to identify \emph{difficult-to-manufacture} features at multiple spatial scales leading to a real-time decision support system for design for manufacturability.
%As a proof of concept, we demonstrate the ability of the framework to identify manufacturability of drilled holes.
\end{abstract}

%%%%%%%%% BODY TEXT

\begin{figure*}[!t]
  \centering
  \includegraphics[width=0.9\textwidth]{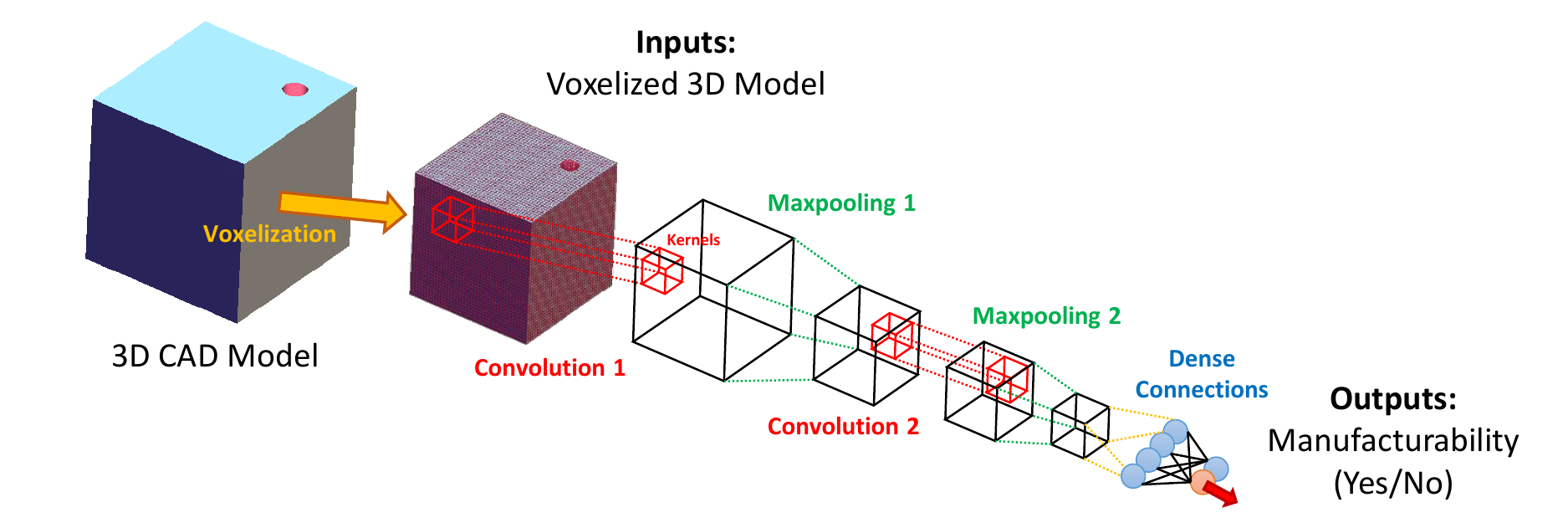}
  \caption{3D convolutional neural network for the classification of whether or not a design is manufacturable. In this example, a block with a drilled hole with specific diameter and depth is considered.}
  \label{Fig:3dcnn}
\end{figure*}

\section{Introduction}
Deep learning (DL) algorithms are designed to hierarchically learn multiple levels of abstractions of the data, and have been extensively used in computer vision~\citep{SVRG14,lee2009convolutional,lore2015llnet,HY08}. More specifically, 3D-Convolutional Neural Networks (3D-CNN) have been used extensively for object recognition, point cloud labeling, video analytics, and human gesture recognition~\citep{maturana20153d,maturana2015voxnet,riegler2016octnet,huang2016point}. Object recognition has been one of the most challenging problems in the area of computer vision and pattern recognition. Early DL-based approaches used simple projection of the 3-dimensional object to a 2-dimensional representation such as depth images or multiple views to recognize an object. However, there is a significant loss of geometric information while using a 2D representation of a 3D object. Therefore, it is difficult to learn about the local features of the object using these projection-based techniques. In this paper, we make use of voxelized 3D representation of the object augmented with surface normal information to identify localized features.

Voxelized models have been successfully used in the past for object recognition. Generally, point cloud data is converted to a volume occupancy grid or voxels to represent the model and identify the object. However, voxel-based occupancy grid representation does not inherently have information regarding the surfaces of the object without additional processing. It is also not robust enough to capture information about the location, size, or shape of a feature within an object. Providing additional information about the geometry of the object has been shown to increase the accuracy of object detection by~\citet{wang2016differential}. In this paper, we propose to use normal information of the surface of the object, in addition to the volume occupancy information, to efficiently learn the local features.

Learning local features in a geometry is different from object recognition. Object recognition is a classification problem, where the object is classified based on a collection of features identified in the object. In this work, we make use of a semi-supervised methodology to learn local features by the DL network, without any specific metric to classify the features. However, we learn localized geometric features of interest within the object based on the cost function of the overall object classification problem. For this purpose, we train a 3D-CNN to learn the key features of the object and also learn the variation in the features that can classify the object for a given cost function.

One of the applications of the aforementioned methodology, which we explore in this paper, is to identify \textit{difficult-to-manufacture} features in a CAD geometry and ultimately, classify its manufacturability. A successful part or a product needs to meet its specifications, while also being feasible to manufacture. In general, the design engineer ensures that the product is able to function according to the specified requirements and the manufacturing engineer gives feedback to the design engineer about its manufacturability. This process is performed iteratively, often leading to longer product development times and higher costs. There are various handcrafted design for manufacturability (DFM) rules that have been used by the design and manufacturing community to ensure manufacturability of the design based on the manufacturing technology and tools available. However, designing a manufacturable product, with fewer design iterations, depends mainly on the expertise of the design and manufacturing engineer.

Conventional statistical machine learning (ML) models require a lot of hand crafting of the features, which make the DFM problem intractable, since there are many complex manufacturing criteria and rules that need to be satisfied for determining manufacturability~\citep{shukor2009manufacturability}. In contrast to this, the hierarchical architecture of DL can be used to learn increasingly complex features by capturing \emph{localized geometric features} and \emph{feature-of-features}. Thus, a deep-learning-based design for manufacturing (DLDFM) tool can be used to learn the various DFM rules from different examples of manufacturable and non-manufacturable components without explicit handcrafting. In addition, the \emph{learned} ML model can be integrated in the CAD system, providing interactive feedback about the manufacturability of the component.

In this paper, we present a 3D Convolutional Neural Network (3D-CNN) based framework that will learn and identify localized geometric features from an expert database in a semi-supervised manner. Further, this is applied to the context of manufacturability with various CAD models classified as \emph{manufacturable} and \emph{non-manufacturable} parts. The main contributions of this paper include:
\begin{itemize}
  \setlength\itemsep{0.0em}
  \item GPU-accelerated methods for converting CAD models to volume representations (voxelization augmented with surface normals), which can be used to learn localized geometric features.
  \item A novel voxel-wise 3D gradient-weighted feature localization based on the 3D-CNN framework to identify local features.
  \item Application of the method to manufacturability analysis of drilled holes.

\end{itemize}

This paper is arranged as follows. We explain the volume representations that we use for 3D-CNN in Section~\ref{Sec:VolumeRep}. In Section~\ref{Sec:3DCNN}, we discuss the details of the 3D-CNN, including the network architecture and the hyper-parameters. In Section~\ref{Sec:GradCAM}, we explain the 3D gradient weighted class activation mapping for identifying the localized geometric features. In Section~\ref{Sec:DFMRules}, we discuss the design for manufacturability (DFM) rules used for generation of the datasets for training and testing the 3D-CNN. Finally, in Section~\ref{Sec:Results}, we show the results of the deep learning based design for manufacturability (DLDFM) framework in classifying \textit{manufacturable} and \textit{non-manufacturable} features and learning capability of the model to identify localized geometric features.

\comment	%%%%%%% Previous section 2 removed
{
\section{Design for Manufacturing Rules}
\label{Sec:DFM}
The objective of the design process is to design products that meet the specifications of the functional requirements, while being able to be manufactured in a cost-efficient manner. Often, the manufacturing process and parameter selection is an optimization problem~\citep{gupta2015optimisation}. Thus, understanding the constraints for the design space requires more information about the manufacturing processes, which makes the concurrent engineering process difficult and manual. 

DFM guidelines were developed by design and manufacturing researchers to make the design process compliant with manufacturing~\citep{BoothroydDewhurst}; however, the application of these guidelines were done manually. In addition, researchers used these DFM guidelines to develop manufacturability analysis systems (MAS), which take into account the rules provided by the DFM guidelines to analyze the manufacturability of the part~\citep{shukor2009manufacturability}. Many MAS frameworks require additional user input regarding different parameters of the part to analyze the manufacturability of the part. These MAS frameworks then use the knowledge base of different DFM rules to provide manufacturability feedback. In addition, there are very few interactive parametric 3D solid modeling tools that provide manufacturability feedback to the designer, and enable changes to the design for improving manufacturability~\citep{shugrina2015fab}.

A common drawback of such tools is that the manufacturability feedback is based on the knowledge base. However, an expert is still needed to make decisions on the part design based on their experience and their understanding of the manufacturing processes. In this paper, we demonstrate the utility of a DLDFM tool on interactively analyzing the manufacturability of a product design without any prior knowledge about the manufacturing processes, with an example of the drilling operation.

Drilling operations are used for manufacturing round holes or to enlarge an existing hole with the help of a drill. The general applications of machined holes may be for bearing a shaft or an axle, threading, fastening, lubrication, porting hydraulic or electric connections, etc.

The various parameters taken into account for deciding the manufacturability of drilled holes are the diameter the hole, the depth of the hole, the material of the part, and the material of the drill tool used. In addition, parameters such as the tolerances of the hole and the accuracy achievable by the drilling machine are also considered. The rule-based principle that relies only on the geometry of the part is based on the depth-to-diameter ratio of the hole to be machined. As the depth of the hole increases, it becomes more and more difficult to maintain the accuracy of the hole. Also, at higher depths with the same diameter, it is possible for the tool to experience high loads sufficient enough to cause damage to the tool. For building the DLDFM model, we assume that a ratio of $10.0$ to be manufacturable~\citep{boothroyd1987product}. Thus, any hole that has the depth-to-diameter ratio of greater than $10.0$ is classified as non-manufacturable. 

Various samples are generated using a CAD modeling kernel, ACIS~\citep{ACIS10}, which is a commercial CAD modeling software that has APIs to create solid models. A cubical block of edge length $5.0$ inches with various sizes of drilled holes is created (Figure~\ref{Fig:sample_model}). The diameter of the hole is varied from $0.1$ in. to $0.5$ in. with an increment of $0.05$ in. Similarly, the depth of the hole is varied from $0.5$ in. to $5.0$ in. with an increment of $0.5$ in. The holes are generated at various positions by varying the value of $PosX$ and $PosY$ (Figure \ref{Fig:sample_model}). In addition, the holes are generated in all the six faces of the cube.

After the CAD models are generated using the solid modeling kernel, they are classified for manufacturability. We use the above mentioned rule for classifying the holes. However, this rule is not completely sufficient to determine the manufacturability of a hole. We take two particular constraints imposed by the human prior: (1) if the hole is a through hole, then the rule shall be that any hole which has the depth to diameter ratio to be greater than $20.0$ will be classified as non-manufacturable. (2) If thickness of the material near the the hole is less than $0.25$ in., then the side wall will be too weak and hence, cannot be manufacturable. Using the CAD models and the classification based on both the rule and the constraints, we develop the DLDFM framework.
}

\section{Volumetric Representations for Learning Geometric Features}
\label{Sec:VolumeRep}

Traditional CAD systems use boundary representations (B-Reps) to define and represent the CAD model ~\citep{Krishnamurthy-2009-CAD}. In B-Reps, the geometry is defined using a set of faces that form the boundary of the solid object. B-Reps are ideally suited for displaying the CAD model by first tessellating the surfaces into triangles and using the GPU to render them. However, learning spatial features using B-Rep can be challenging, since the B-Rep does not contain any volumetric information.
\subsection{Voxel based representation of Geometry}
In our framework, we convert the B-Rep CAD model to a volumetric occupancy grid of voxels. However, voxelizing a B-Rep CAD model is a compute intensive operation, since the center of each voxel has to be classified as belonging to either inside or outside the model. In addition, thousands of models need to be voxelized during training. Traditional CPU voxelization algorithms are too computationally slow for training the machine learning network in a reasonable time frame. Hence, we have developed methods for accelerated voxelization of CAD models using the graphics processing unit (GPU). These GPU methods are more than 100x faster than the existing state-of-the-art CPU-based methods and can create a voxelized representation of the CAD model with more than 1,000,000,000 voxels. Having a high resolution voxelization will enable us to capture small features in a complex CAD model.

\begin{figure}[b]
  \centering
  \includegraphics[width=2.0in]{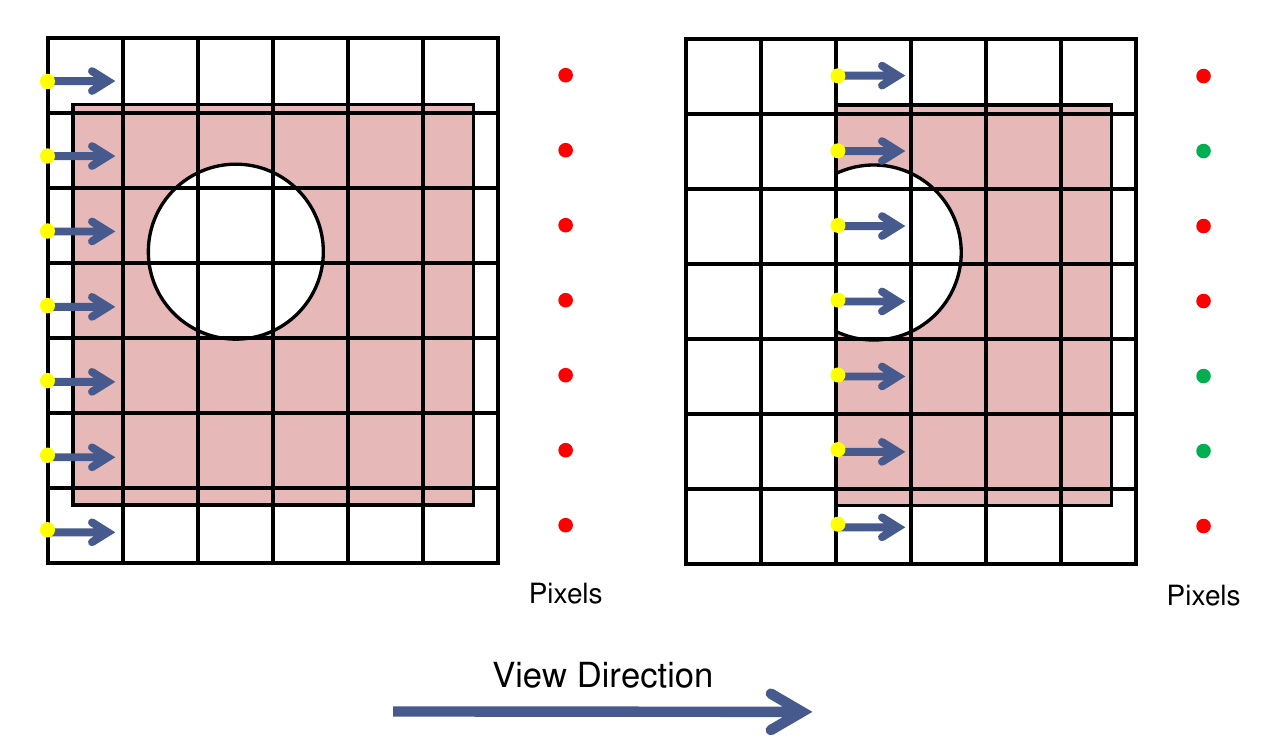}
  \caption{Performing voxelization in 2D using GPU rendering. A clipped CAD model is rendered slice-by-slice and the the number of rendered pixels is counted. The pixels that are rendered an odd number of times in each slice are inside the object (green).}
  \label{Fig:Voxelization2D}
\end{figure}

\begin{figure}[!t]
  \centering
  \includegraphics[width=3.0in]{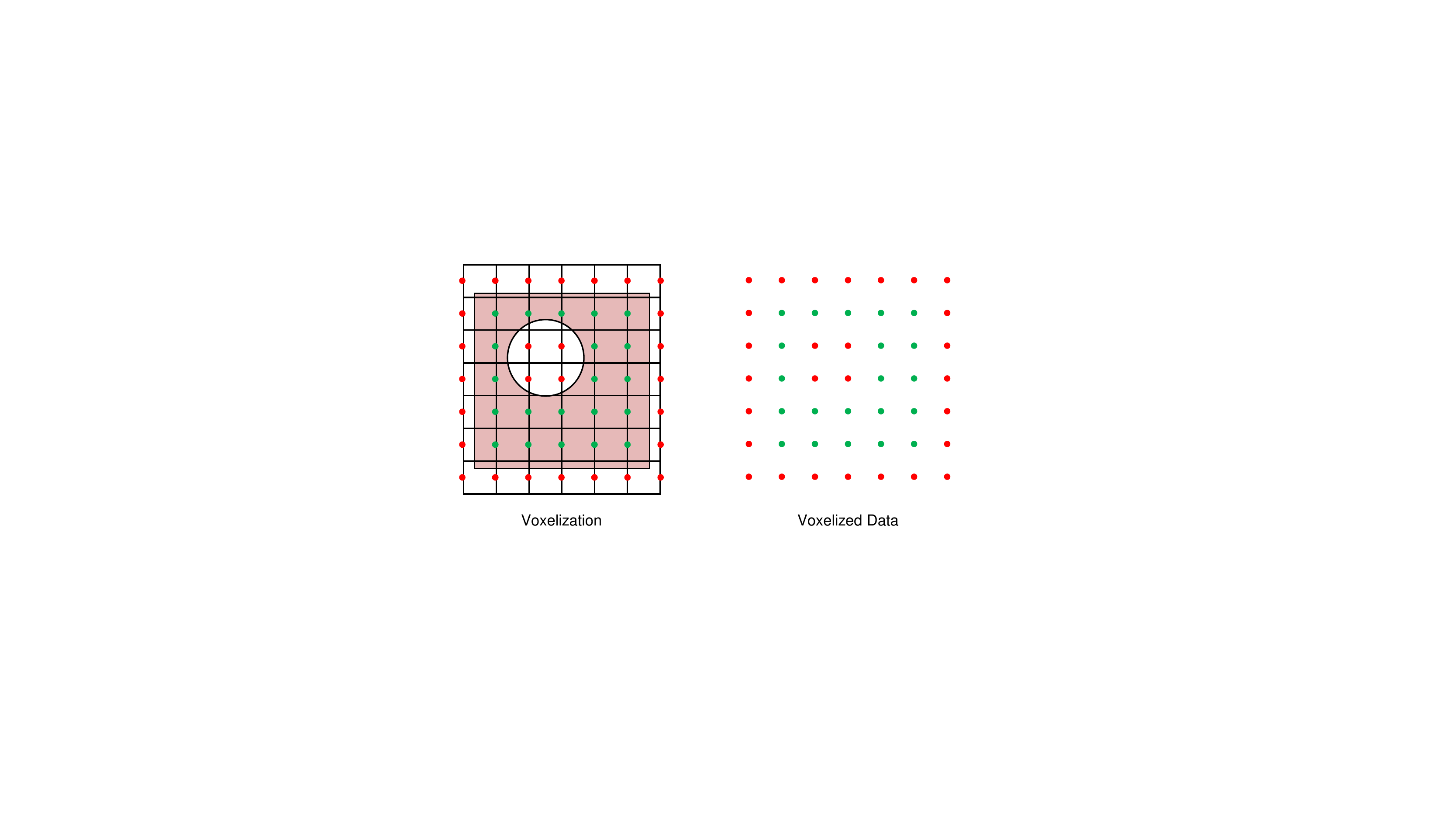}
  \caption{Visualization of voxelized CAD models. The voxels marked in red are outside the model while the voxels marked in green are inside.}
  \label{Fig:Voxelization2DExample}
\end{figure}

\begin{figure}[!b]
  \centering
  \includegraphics[width=2.4in]{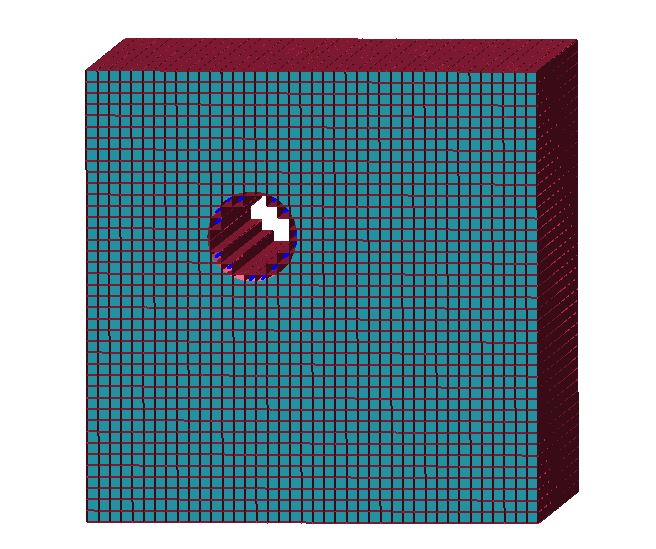}
  \caption{Example voxelized 3D CAD model. The voxels that make up the model are rendered as boxes.}
  \label{Fig:Vol_Surf}
\end{figure}

To create the voxelized CAD model, we construct a grid of voxels in the region occupied by the object. We then make use of a rendering-based approach to classify the voxel centers as being inside or outside the object. A 2D example of the method is shown in Figure~\ref{Fig:Voxelization2D}; the method directly extends to 3D. The CAD model is rendered slice-by-slice by clipping it while rendering. Each pixel of this clipped model is then used to classify the voxel corresponding to the slice as being inside or outside the CAD model. This is performed by counting the number of fragments that were rendered in each pixel using the stencil buffer on the GPU. After the clipped model has been rendered, an odd value in the stencil buffer indicates that the voxel on the particular slice is inside the CAD model, and vice versa (Figure~\ref{Fig:Voxelization2DExample}). The process is then repeated by clipping the model with a plane that is offset by the voxel size. Once all the slices have been classified, we get the complete voxelized representation of the CAD model (Figure~\ref{Fig:Vol_Surf}).

%Using this method on the GPU, a fine voxelization of the model (up to 1 billion voxels) with a relative voxel size of 0.001, can be generated (Figure~\ref{Fig:Vol_Surf}). This resolution is fine enough to create a voxelization that can recreate the fine features in a CAD model.

%We make use of the method developed by \citet{Krishnamurthy-2009-CAD} to directly evaluate and render the NURBS surfaces in the model using the GPU. The B-Rep model is first decomposed into its component surfaces. If the surface is a flat surface, it is converted into triangles with a very fine resolution that is less than one-tenth of the voxelization resolution. All other surfaces are converted into NURBS and are evaluated using the GPU while rendering. Each surface is rendered successively to an off-screen framebuffer and the stencil buffer is used to classify the voxels corresponding to the slice as being inside or outside as explained above. Hence, the model is rendered once for each slice to classify the voxelization. 

The time taken to perform the classification is the sum of the time taken to tessellate the model once and the total time taken to render each slice. As an example, the total time taken to voxelize the hole block is 0.133 seconds. These timings are obtained by running our voxelization algorithm on a Intel Xeon CPU with 2.4 GHz processor, 64 GB RAM, and an NVIDIA Titan X GPU.

\subsection{Augmentation of Surface Normals to volume representation}
While a digital voxelized representation may be sufficient for a regular object recognition problem, it may not be enough to capture the detail geometrical information for identifying local fetaures of interest within the object. In particular, information about the boundary is lost in a voxel grid and the local region around a voxel grid need to be analyzed to identify a boundary voxel. To overcome this challenge in voxelized representations, we augment the voxel occupancy grid with the surface normals of the B-Rep geometry. First, we identify the boundary voxels of the B-Rep geometry. We consider each voxel as an axis-aligned bounding-box (AABB) using the center location and size of each voxel. We then find all the triangles of the B-Rep model that intersect with the AABBs. Finally, we average the surface normals of all triangles that intersect with each AABB. The $x, y, \& z$ components of the surface normals are then embedded in the voxelization along with the occupancy grid.

The surface normal information can be augmented with the volume occupancy information in each voxel in two distinct ways:
\begin{itemize}
  \setlength\itemsep{0.0em}
  \item The volume occupancy grid and the surface normals are represented independently as four channels (voxel, $X,Y$ and $Z$ directions) for the same object.
  \item It can be noted that in the previous representation, the normals information exists only on the boundary voxels and is zero all other places in the grid. Alternative representation would be to fuse the volume occupancy grid information with the boundary voxels containing the surface normals (making it 3 channels for the same object).  
\end{itemize}

While one representation is more sparse  which might be helpful while training the 3D-CNN, it is also to be noted that more memory is needed while training. In this paper, we explore both the methods of representing the normal information.    

\section{3D-CNN for Learning Localized Geometric Features}
\label{Sec:3DCNN}

The voxel based representation with the surface normals of the CAD model can be used to train a 3D-CNN that can identify local features. 3D-CNNs have mostly been used for complete 3D object recognition and object generation problems~\citep{maturana20153d,maturana2015voxnet,riegler2016octnet,riegler2017octnetfusion,tatarchenko2017octree,huang2016point} as well as analyzing temporally correlated video frames~\citep{ji20133d,tran2015learning,payan2015predicting,lee2015recursive}. However, to the best of the authors' knowledge, this is the first application of 3D-CNNs on identifying local features with applications in manufacturability analysis.

\subsection{Network Architecture and Hyper-Parameters}

\label{SubSec:architecture}
The input to the 3D-CNN is a voxelized CAD model. The input volumetric data is first padded with zeros before convolution is performed. Zero padding is necessary in this case to ensure that the information about the boundary of the CAD model is not lost while performing the convolution. The convolution layer is applied with RELU activation. The convolutional layer is followed by batch normalization layer and Max. Pooling layer. The same sequence of Convolution, batch normalization and Max. Pooling is again used. A fully connected layer is used before the final output layer with sigmoid activation. The hyper-parameters of the 3D-CNN that needs to be tuned in order to ensure optimal learning. The specific hyper-parameters used in our framework are listed in Section~\ref{Sec:Results}.

The model parameters $\theta$, comprised of weights $\textbf{W}$ and biases, $\textbf{b}$ are optimized by error back-propagation with binary cross-entropy as the loss function~\citep{hinton2006reducing} using the ADADELTA optimizer~\citep{zeiler2012adadelta}. Specifically, the loss function $\ell$ to be minimized is:
\begin{equation}\label{eqn:bce}
\ell = -y\log {\hat  {y}}-(1-y)\log(1-{\hat  {y}})
\end{equation}
\noindent where $y\in \{0,1\}$ is the true class label and $\hat{y} \in \{0,1\}$ is the class prediction.

%\begin{figure*}[!t]
%  \centering
%  \includegraphics[width=1.0\textwidth, clip, trim={0.0in 1.7in 0.5in 0.7in}]{Feature_Maps.pdf}
%  \caption{Feature map visualization with respect to the second (a, b, c) and the third (d, e, f) layers of the 3D-CNN trained using %\emph{in-outs} information (top row) and orthogonal distance fields (bottom row).}
%  \label{Fig:fminouts}
%\end{figure*}

For training the network, many CAD models were generated based on the method illustrated in section \ref{Sec:DFMRules}.

\section{Interpretation of 3D-CNN Output}
\label{Sec:GradCAM}
The trained 3D-CNN network can be used to classify the object of any new geometry and can be treated as a black-box. However, interpretability and explainability of the output provided by the 3D-CNN are very essential. In this paper, we attempt to visualize the input features that lead to a particular output and if possible modify it. A similar approach was used in object recognition in images by using class activation maps to obtain class specific feature maps~\citep{selvaraju2016grad}. The class specific feature maps could be obtained by taking a class discriminative gradient of the prediction with respect to the feature map to get the class activation. In this paper, we present the first application of gradient weighted class activation map (3D-GradCAM) for 3D object recognition.

In order to get the feature localization map using 3D-GradCAM, we need to compute the spatial importance of each feature map $A_l$ in the last convolutional layer of the 3D-CNN, for a particular class, $c$ ($c$ can be either non-manufacturability or manufacturability, for the sake of generality) in the classification problem. This spatial importance for each feature map can be interpreted as weights for each feature map; it can be computed as the global average pooling of the gradients back from the specific class of interest as shown in Eqn.~\ref{Eqn:GlobalAvgPooling}.

The cumulative spatial importance activations that contribute to the class discriminative localization map, $L_{3DGradCAM}$, is computed using
\begin{equation}
\label{Eqn:GradCAM}
L_{3DGradCAM} = ReLU\left(\sum_l{\alpha_{l} \times A^l}\right),
\end{equation}
\noindent where $\alpha_{l}$ are the weights computed using
\begin{equation}
\label{Eqn:GlobalAvgPooling}
\alpha_{l} = \frac{1}{Z} \times \sum_i{ \sum_j{ \sum_k{ \frac{\partial y^c}{\partial A_{ijk}^l}}}}.
\end{equation}
\noindent We can compute the activations obtained for the input part using $L_{3DGradCAM}$ to analyze the source of output. The heat map of $\left(L_{3DGradCAM}\right)$ is resampled using linear interpolation to match the input size, and then overlaid in 3D with the input to be able to spatially identify the source of non-manufacturability. This composite data is finally rendered using a volume renderer.

We make use of a GPU-based ray-marching approach to render this data. The rendering is parallelized on the GPU with each ray corresponding to the screen pixel being cast independently. The intersection of the ray with the bounding-cube of the volumetric data is computed, and then the 3D volumetric data is sampled at periodic intervals. The sum of all the sampled values along the ray is then computed. This value is converted to RGB using a suitable color-bar and rendered on the screen. Table~\ref{Tab:GradCAM} shows different volumetric renderings of the composite 3D-GradCAM data.

\section{Manufacturability of Drilled Holes}
\label{Sec:DFMRules}
Design for Manufacturing (DFM) rules for drilling have been traditionally developed based on the parameters of the cylindrical geometry as well as the geometry of the raw material. In this paper, we show an application of our approach to learn localized geometric features to identify non-manufacturability of drilled holes. Deciding the manufacturability of a part is framed as a binary classification problem. We apply the above methodology to learn the parameters that correspond to a hole feature, and then use the learned parameters to classify for manufacturability.

The important geometric parameters of a hole are the diameter, the depth, and the position of the hole. However, there are certain additional geometric parameters that do not contribute to the manufacturability analysis, but might affect the machine learning framework. For example, the face of the stock on which the hole is to be drilled does not affect the manufacturability of the hole, but need to be considered while training because the volumetric representation of the CAD model is not rotationally invariant.

In our DLDFM framework, the following DFM rules are used to classify the drilled hole as manufacturable.
\begin{enumerate}
 	\setlength\itemsep{0.0em}
	\item \textbf{Depth-to-diameter ratio}: The depth-to-diameter ratio should be less than $5.0$ for the machinability of the hole~\citep{BoothroydDewhurst,bralla1999design}. It should be noted that this rule is generic and applicable for all materials.
    \item \textbf{Through holes}: Since a through hole can be drilled from both directions, the depth-to-diameter ratio for a through hole should be less than $10.0$ to be manufacturable.
    \item \textbf{Holes close to the edges}: A manufacturable hole should be surrounded with material of thickness at least equal to the half the diameter of the hole.
    \item \textbf{Thin sections in the depth direction of the hole}: A manufacturable hole should should have material greater than half the diameter along the depth direction. 
\end{enumerate}

The preceding rules are used to generate the ground truth manufacturability data for the training set, which is then used to learn the manufacturable and non-manufacturable features by the DLDFM Framework. However, it should be noted that for a DLDFM framework, one need not explicitly mention the rule. Rather, for industrial applications, one can train the DLDFM framework using the industry relevant historical data available in the organization, which need not be strictly rule based. The historical data can also be based on experience during previous attempts to manufacture a part. Thus, this DLDFM framework is an attempt to generate a DFM framework based on few basic local features. For more complicated shapes, one can just augment the training set and then re-train the DLDFM framework, which can then be used for analyzing complex parts for manufacturability. This eliminates the hand-crafting of rules, ultimately leading to better manufacturability analysis.

\subsection{Training Data}
\label{SubSec:TrainingData}

\begin{figure}[!t]
  \centering
  \includegraphics[width=3.0in]{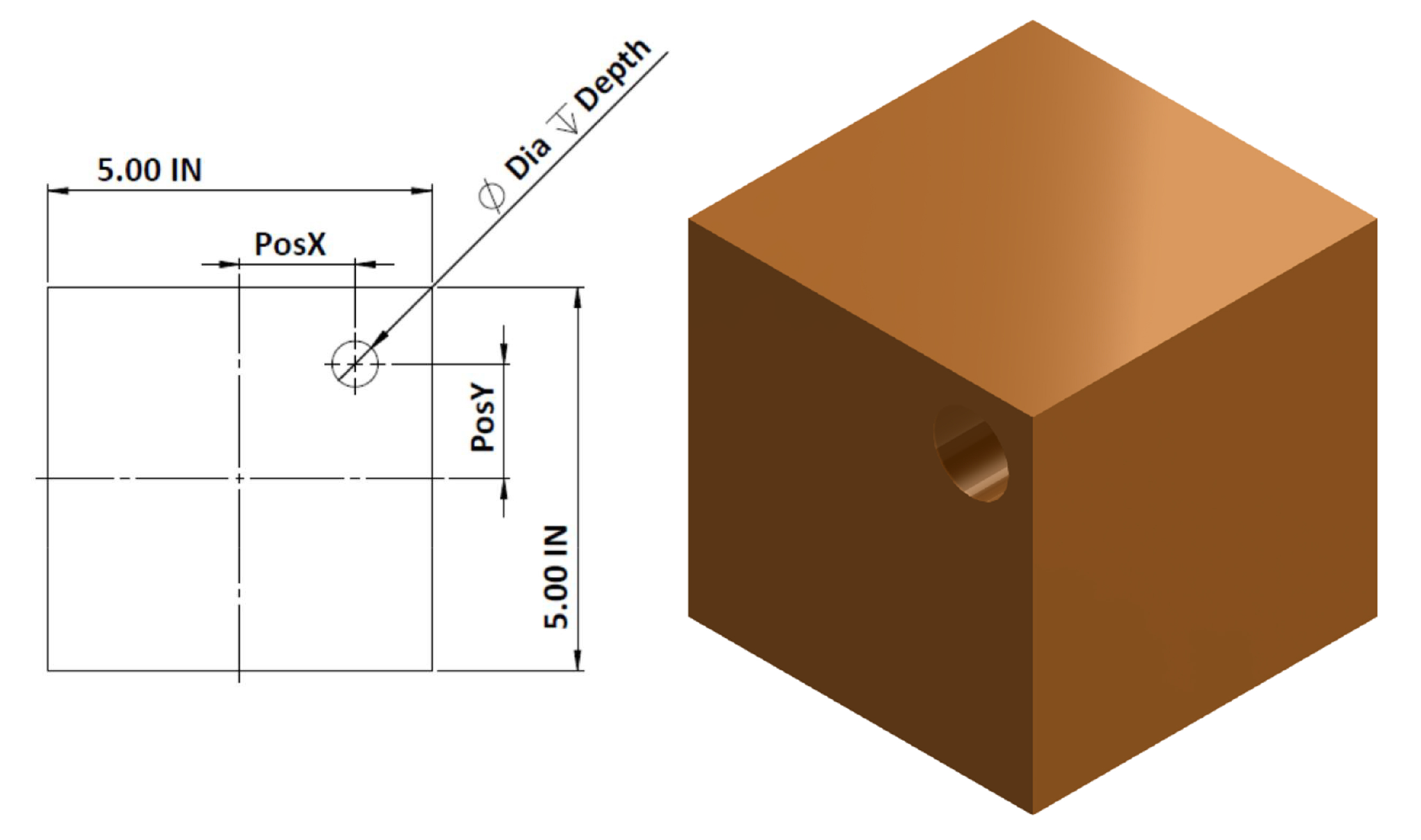}
  \caption{A sample block with a drilled hole with its dimensions highlighted in the projected view.}
  \label{Fig:sample_model}
\end{figure}

Based on the DFM rules for drilling, different sample solid models are generated using a CAD modeling kernel. We use ACIS~\citep{ACIS10}, a commercial CAD modeling kernel to create the solid models. A cubical block of edge length $5.0$ inches with different sizes of drilled holes are created (Figure~\ref{Fig:sample_model}). The diameter of the hole is varied from $0.1$ in. to $1.0$ in. with an increment of $0.1$ in. Similarly, the depth of the hole is varied from $0.5$ in. to $5.0$ in. with an increment of $0.5$ in. In addition, few geometries are specifically generated with thin sections in the depth direction to make the training data complete. The holes are generated at different positions on the face of the cube by varying the value of $PosX$ and $PosY$ (Figure \ref{Fig:sample_model}). For sake of simplicity, the holes are located only along the diagonal of the drilling face, i.e. $PosX=PosY$ for all the training samples generated. In addition, the holes are generated in all the six faces of the cube. After the CAD models are generated using the solid modeling kernel, they are classified for manufacturability using the DFM rules for drilled holes.

After the B-Rep models are generated using the CAD modeling kernel, they are converted to volume representations. One of the framework design choices is to choose an appropriate voxel grid for the model. A fine voxelization of the model, while capturing all the features accurately, might be computationally expensive for training the 3D-CNN. Even a voxelization resolution of $64 \times 64 \times 64$ pushes the limits of the GPU and CPU memory, and hence, the parameters of the 3D-CNN have to be tuned for optimal performance. The complete data is split into training set and validation set.

%While the data is available in terms of B-Rep, for training the DLDFM model one of the major concerns is the resolution of the volumetric representation of the models. One might use a very small refinement of the grid, but shall end up computationally intensive. Even for training a $64 \times 64 \times 64$ grid volumetric representation, one might face issues while training the DLDFM model as the large data with fine resolution  takes a lot of memory in both GPU and CPU. Thus, it is extremely important to choose the right resolution of the volumetric grid in order to capture the features accurately while not compromising on the performance of the DLDFM framework.

\subsection{Representative and Non-Representative Data}
\label{SubSec:Models}

In order to test the performance of the DLDFM network, a new test set of CAD models are generated. The CAD geometry generated in this set is different from the CAD models used in training the DLDFM. Further, we split the test set into two classes, representative and non-representative.

Representative models are broadly defined as those geometries that are related to CAD models in the training set. For example, the DLDFM network was trained using models with a single hole on different faces and at different positions. The geometries in the test set that belong to this subset of the CAD model parameters are classified as representative geometries. However, the models generated in the representative set do not have the same depth or diameter values as those in the models of the training set. For the representative test data, the following parameters are varied to generate the samples.

\begin{itemize}
  \setlength\itemsep{0.0em}
  \item Diameter values from $1.1$ to $1.5$ are used for the representative test data. Diameter values from $0.1$ to $1.0$ inches were used for training.
  \item Position of the holes is varied in the radial directions ($PosX=0$ or $PosY=0$, while the other is varied), as well as in the diagonal direction. The position of the hole vary only in the diagonal direction ($PosX=PoxY$) in the training samples.
\end{itemize}

Non-representative models are broadly defined as those geometries that are completely different from the training set. The generalization ability of DLDFM can be tested by creating a non-representative test set, containing geometries with the same primary hole parameters (depth, diameter, and position of the hole), but having additional or different external features. The details of the geometries in the non-representative data set is given below.

\noindent \textbf{Multiple holes}: The DLDFM has been trained to analyze the manufacturability of a single drilled hole. However, in an a designed component, the features may not be independent; there can be multiple features, each of which may or may not be manufacturable. Moreover, it is possible that each of the features themselves are manufacturable, but due to their proximity or interaction with other features, the part may become non-manufacturable. Hence, we test the ability of the DLDFM framework to analyze the manufacturability of a part with two holes.

\noindent \textbf{L-shaped blocks and cylindrical models}: All the models in the training set have an external cubical shape. Hence, to test the capability of DLDFM to capture the manufacturability of a hole irrespective of the external geometry, we use L-shaped Block and cylinders with holes. The rules established in Section~\ref{Sec:DFMRules} also apply to this geometry.

We generated 9531 CAD models in total for the training and validation set. Out of these, 75\% of the models were used for training the 3D-CNN and the remaining 25\% of the models were used for validation or fine-tuning the hyper-parameters of the 3D-CNN. A detailed description of the training process is provided in Section~\ref{Sec:3DCNN}. The trained DLDFM network is then tested using the test set CAD models. The test set contains 675 representative geometries and 1450 non-representative geometries.

\begin{table*}[!t]

  \centering
  \small
  \newcommand\T{\rule{0pt}{2.7ex}}
  \newcommand\B{\rule[-1.3ex]{0pt}{0pt}}
  \newcommand{\tabincell}[2]{\begin{tabular}{@{}#1@{}}#2\end{tabular}}
  \tymin=.1in
  \tymax=2.5in 
  \begin{tabulary}{7.0in}{|L|C|R|R|R|R|R|}
    \hline
    Test Data Type&\tabincell{l}{Model Description\B}&\tabincell{l}{True \T\\ Positive\B}&\tabincell{l}{True \T\\ Negative\B}&\tabincell{l}{False \T\\ Positive\B}&\tabincell{l}{False \T\\ Negative\B}&\tabincell{l}{Accuracy\B}\\   
    \hline                
    \multirow{2}{*}{\tabincell{l}{Representative \\675 models\\408 Manufacturable}}& In-outs Information\T\B& 391& 90 & 17 & 176& 0.7136\\
    \cline{2-7}
    &In-outs + Surface Normals \T\B& 334& 201 & 74 & 65& 0.7938\\
    \cline{2-7}
    &Coupled In-outs with Surface Normals \T\B& 405& 131& 3& 135& \textbf{0.7952}\\
    \hline
    \multirow{2}{*}{\tabincell{l}{Non-Representative \\1450 models\\724 Manufacturable}}&In-outs Information\T\B& 500& 422 & 226 & 301 & 0.6363 \\
    \cline{2-7}
    \tabincell{l}{}&In-outs + Surface Normals\T\B&  356 & 601 & 370 & 122 & \textbf{0.6604}\\
    \cline{2-7}
    \tabincell{l}{}&Coupled In-outs with Surface Normals\T\B&  370 & 582 & 356 & 141 & 0.6570\\
    \hline
  \end{tabulary}
  \caption{Quantitative performance assessment of the DLDFM on representative and non-representative data sets.}
  \label{Tab:conf_test}
\end{table*}

\section{Results and Discussion}
\label{Sec:Results}
The different CAD geometries generated as explained in Section~\ref{SubSec:TrainingData} are classified to be manufacturable or non-manufacturable based on the rules discussed in Section~\ref{Sec:DFMRules}. The B-Rep CAD geometries are converted to volumetric representation using voxelization as explained in the Section~\ref{Sec:VolumeRep}. The grid size of $64 \times 64 \times 64$ is used for the volumetric representation in order to represent the geometry with sufficient resolution. We initially use the voxelized representation of the CAD geometry to train the DLDFM network. Then, we use the surface normal information (i.e. $x, y, z$ components) in addition to the voxelized representation of the CAD geometry as input. These are considered as \textit{four} channels of the 3D-CNN input to train another DLDFM network. Since the normal information exists only at the boundary voxels, the three channels (corresponding to the normal components) are very sparse and hence, might be difficult effectively be used by the 3D-CNN. Hence, we fuse the voxelized representation with each of the surface normals (\textit{coupled normal information}) and provide this as a a three channel input to the 3D-CNN to train the DLDFM network. We set all 3 channels to be 1 inside the object and to be 0 outside the object. The boundary voxels have the value of the 3 channels corresponding to the 3 components of the surface normal.

\subsection{Tuning of the Hyper-Parameters}
\begin{figure}[!t]
  \centering
  \includegraphics[width=\columnwidth]{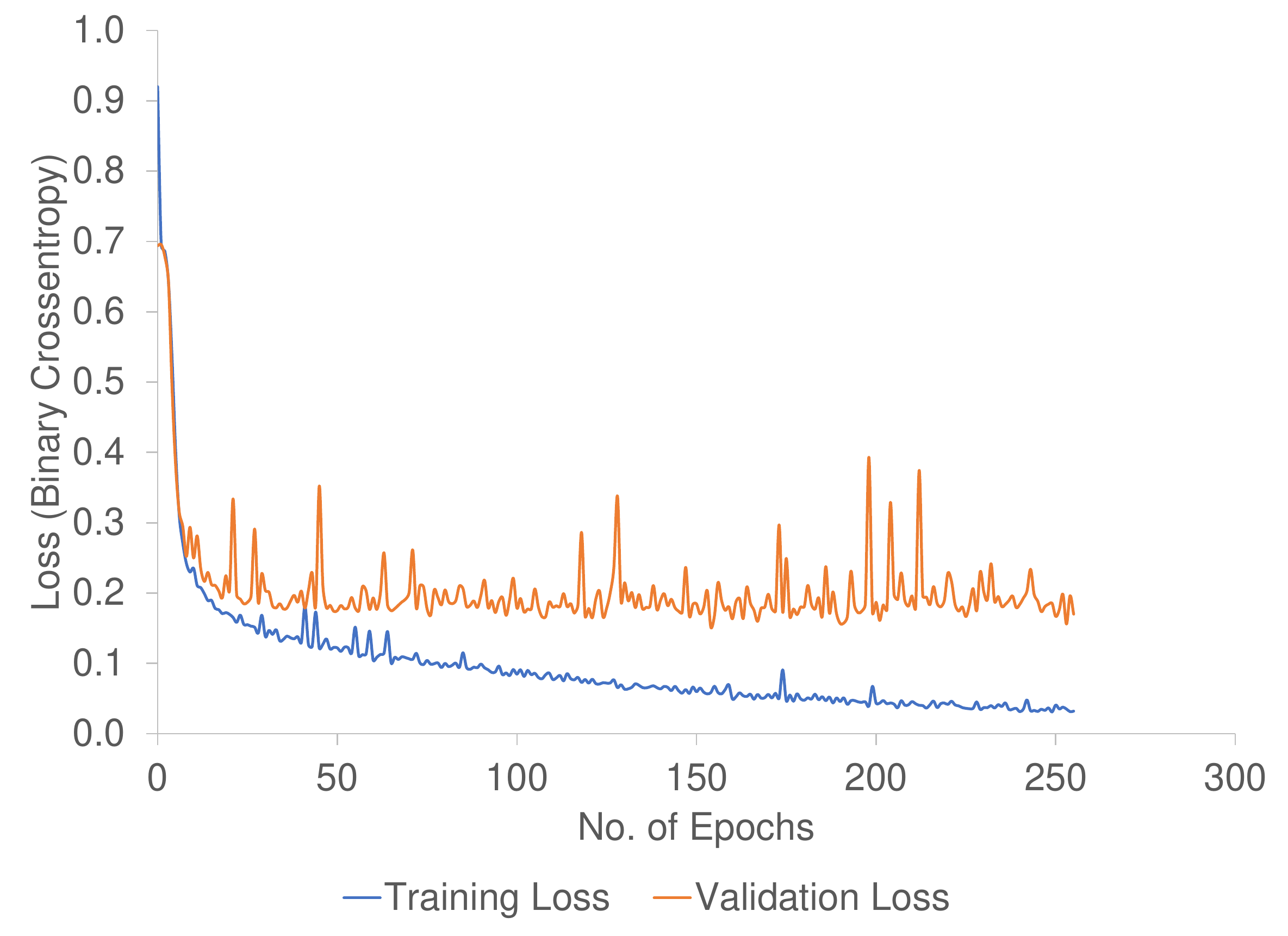}
  \caption{Loss vs. No. of Epochs while training}
  \label{Fig:loss}
\end{figure}

% details of the hyper-parameters and architecture
The hyper-parameters for the DLDFM are fine-tuned to have least validation loss. The architecture of the DLDFM network with voxelized information is composed of three convolution layers with filter sizes of 8, 4, and 2 respectively. Likewise, the DLDFM networks using surface normal information along with voxelized representation, comprises of three convolution layers with filter sizes of 6, 3, and 2 respectively. In succession to the first and last Convolution layers, we use MaxPooling layers of subsampling size 2. A batch size of 64 is selected while training all of the three DLDFM networks. The training was performed using Keras~\citep{chollet2015keras} with a TensorFlow~\citep{tensorflow2015-whitepaper} backend in Python environment. The training of DLDFM networks was performed in a workstation with 128GB CPU RAM and a NVIDIA Titan X GPU with 12GB GPU RAM. The training is run for a lot of epochs and is stopped when the validation loss is not improving for a patience 10 epochs. The training loss vs. validation loss of the DLDFM is shown in Figure~\ref{Fig:loss}.

\subsection{Test Results}

\begin{figure*}[!b]
  \centering
  \includegraphics[width=0.9\textwidth]{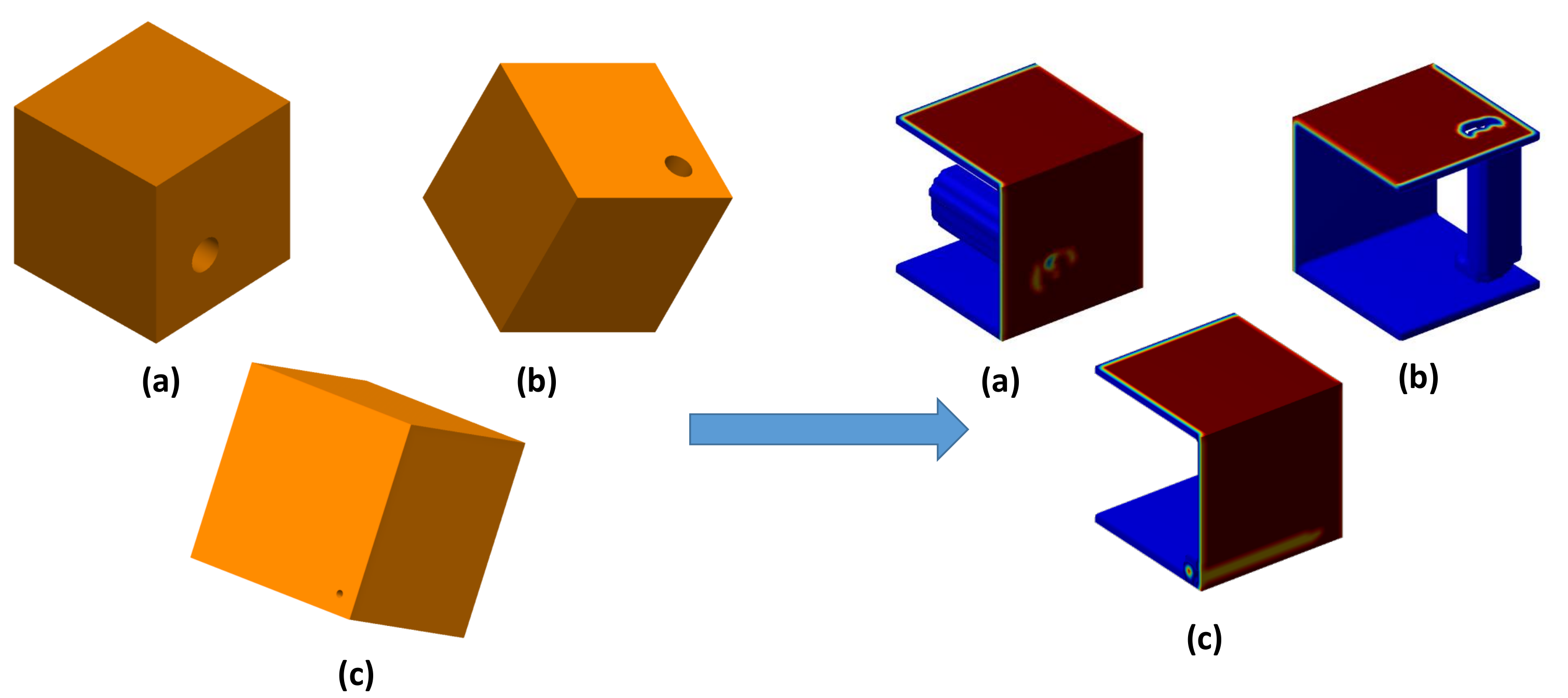}
  \caption{The output of one of the first hidden layer filters in the 3D-CNN architecture. Here, CAD model (a) represents a through hole, CAD model (b) represents a big and deep hole, and CAD model (c) represents a small hole near the corner.}
  \label{Fig:fm_comp}
\end{figure*}

After successful training, the DLDFM network was tested on the representative test-set to benchmark its performance. The \textit{voxel} based DLDFM network performs well with an accuracy of 71\% on 675 models of the representative test set. However, as expected, the performance of the network reduces by a small percentage while testing on the non-representative test set (Table~\ref{Tab:conf_test}). In addition, even though the surface normals augmented DLDFM networks have a poorer training accuracy, their prediction capability on the test-set (both representative and non-representative) is significantly better than the voxel based approach. Specifically, there was a significant improvement in the overall accuracy (a gain of 6\%) for predictions in the representative test set as shown in Table~\ref{Tab:conf_test}. \textit{Note that the test-set has completely different geometries compared to the training set. Thus, it can be seen that the DLDFM is learning the localized geometric features}. 

%In addition, it can been seen that augmenting the surface normal information with voxel data helps in correctly classifying the non-manufacturable features. The main error is contributed by high false-positives, which makes the DLDFM network conservative in classifying manufacturable designs. This is preferable, since borderline manufacturable designs probably need to be modified to account for variations in the manufacturing process itself.

%The feature maps obtained from the DLDFM network is able to recognize the features (Figure~\ref{Fig:fm_comp}). Furthermore, the number of examples that are falsely classified to be manufacturable using the \textit{in-outs} is reduced by using the orthogonal distance fields in both the representative and non-representative test set. 

\begin{table*}[h!]
\newcommand\T{\rule{0pt}{2.6ex}}
\newcommand\B{\rule[-1.2ex]{0pt}{0pt}}
\newcommand{\tabincell}[2]{\begin{tabular}{@{}#1@{}}#2\end{tabular}}
\centering
\tymin=.2in
\tymax=2.0in 
% Figures Table
\begin{tabulary}{7.0in}{|L|C|C|C|L|}
\hline%row 1
\textbf{Feature\T} & \textbf{DFM} & \textbf{DLDFM Prediction\T} & \textbf{CAD Model\T\B} & \tabincell{c}{\textbf{3D-GradCAM\T}\\\textbf{Visualization\B}} \\
\hline
\raisebox{-10mm}{\tabincell{l}{Single hole}}
&
\raisebox{-10mm}{\tabincell{l}{Manufacturable}}
& 
\raisebox{-10mm}{\tabincell{c}{Manufacturable \\ (Feature: Depth/Diameter ratio\\ of hole $<$ 5)}}
& 
\raisebox{-10mm}{\tabincell{l}{\includegraphics[width=20mm, height=20mm]{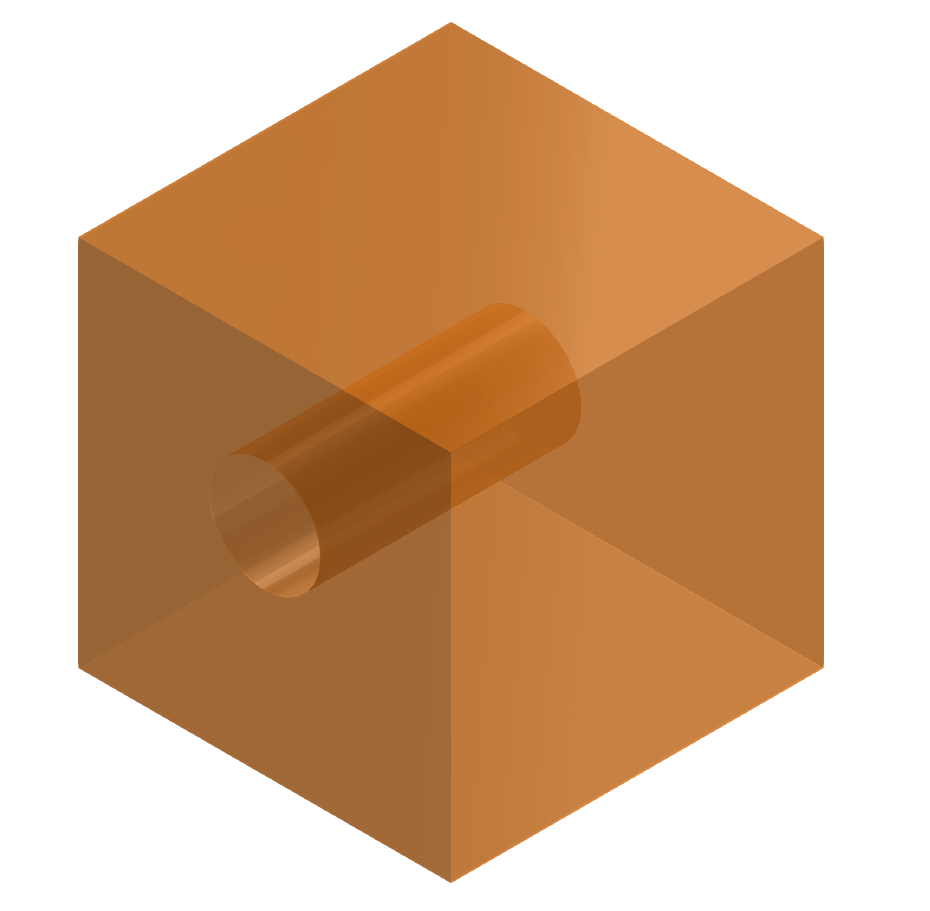}}}
&
\raisebox{-10mm}{\tabincell{l}{\includegraphics[width=20mm, height=20mm]{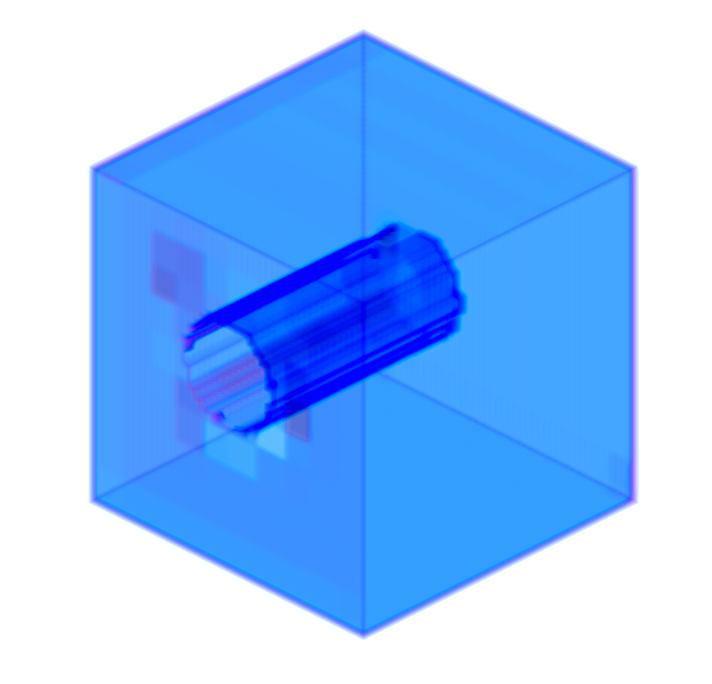}}
\raisebox{-8mm}{\includegraphics[width=10mm, height=25mm]{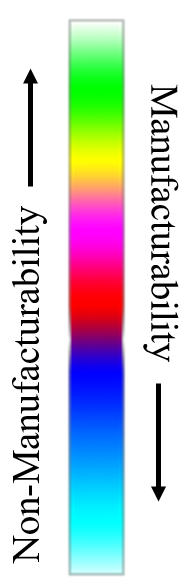}}} 

\\
\hline%row 2
\raisebox{-10mm}{\tabincell{l}{Single hole}}
&
\raisebox{-10mm}{\tabincell{l}{Non-Manufacturable}}
& 
\raisebox{-10mm}{\tabincell{c}{Non-Manufacturable \\ (Feature: Hole close\\ to the edge)}}
& 
\raisebox{-10mm}{\tabincell{l}{\includegraphics[width=20mm, height=20mm]{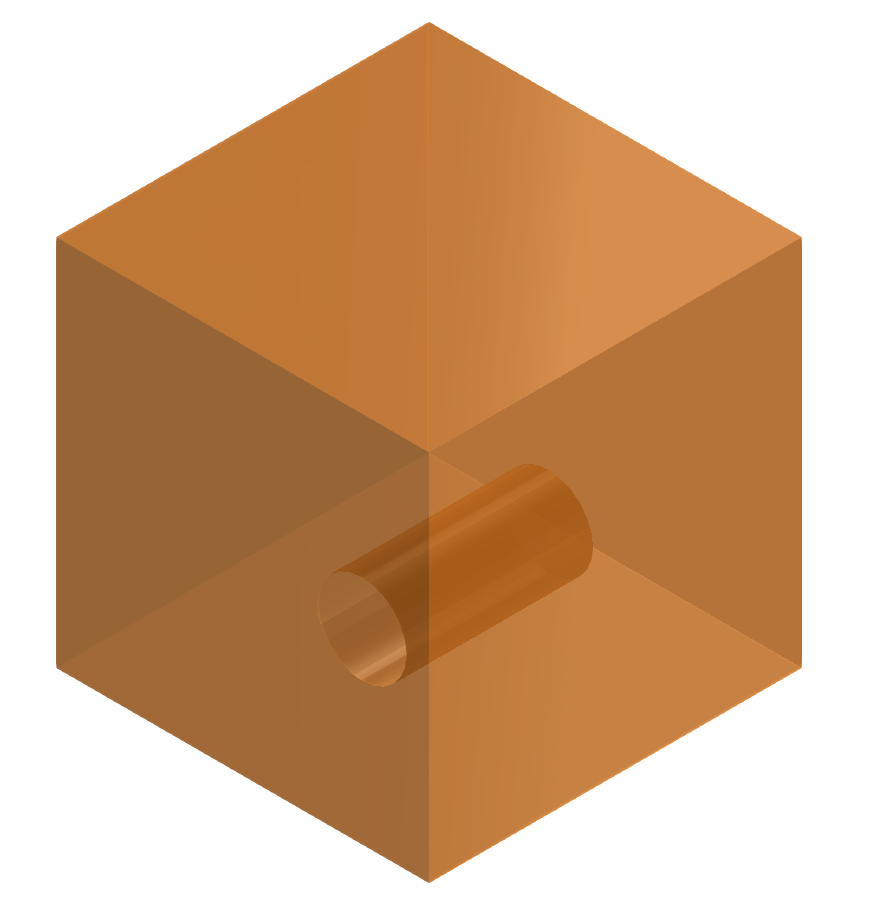}}}
&
\raisebox{-10mm}{\tabincell{l}{\includegraphics[width=20mm, height=20mm]{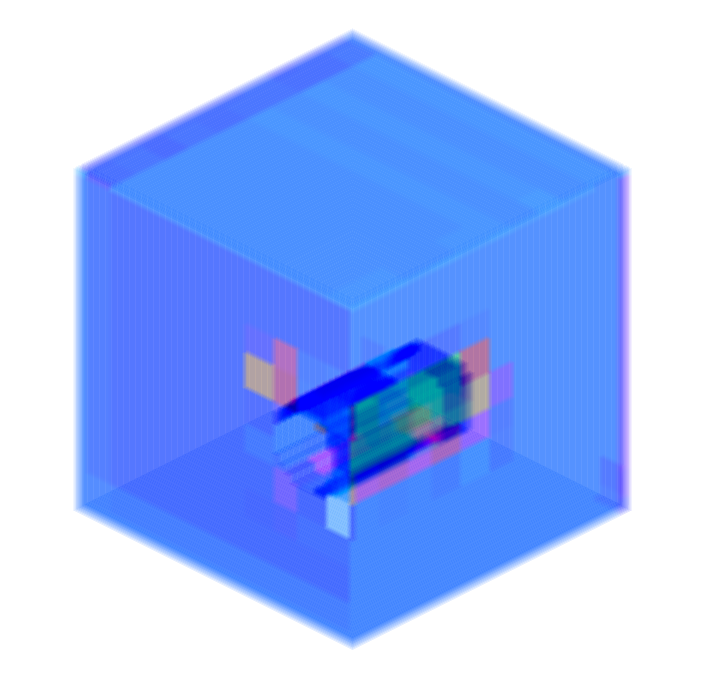}}}

\\
\hline%row 3
\raisebox{-10mm}{\tabincell{l}{Two holes\\Same face}}
&
\raisebox{-10mm}{\tabincell{l}{Non-Manufacturable}}
& 
\raisebox{-10mm}{\tabincell{c}{Non-Manufacturable\\ (Feature: Distance between holes\\ $<$ diameter)}}
& 
\raisebox{-10mm}{\tabincell{l}{\includegraphics[width=20mm, height=20mm]{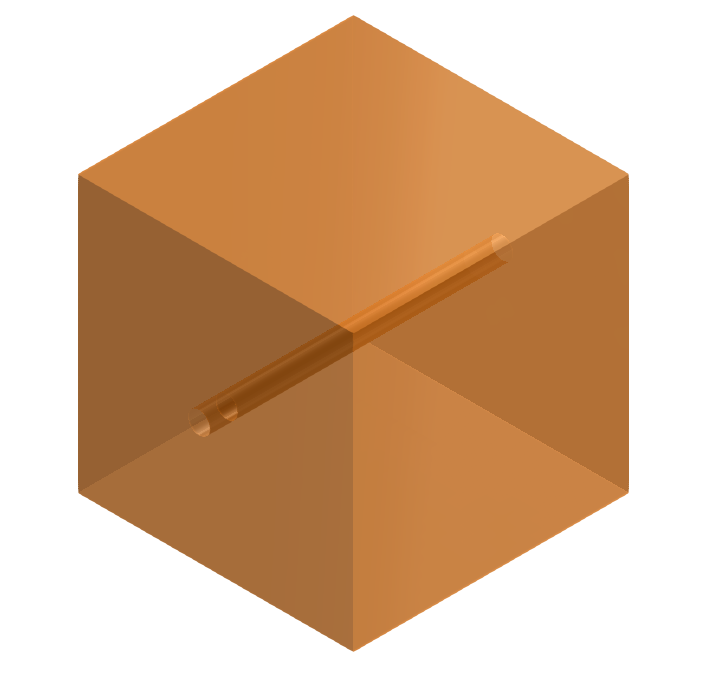}}}
&
\raisebox{-10mm}{\tabincell{l}{\includegraphics[width=20mm, height=20mm]{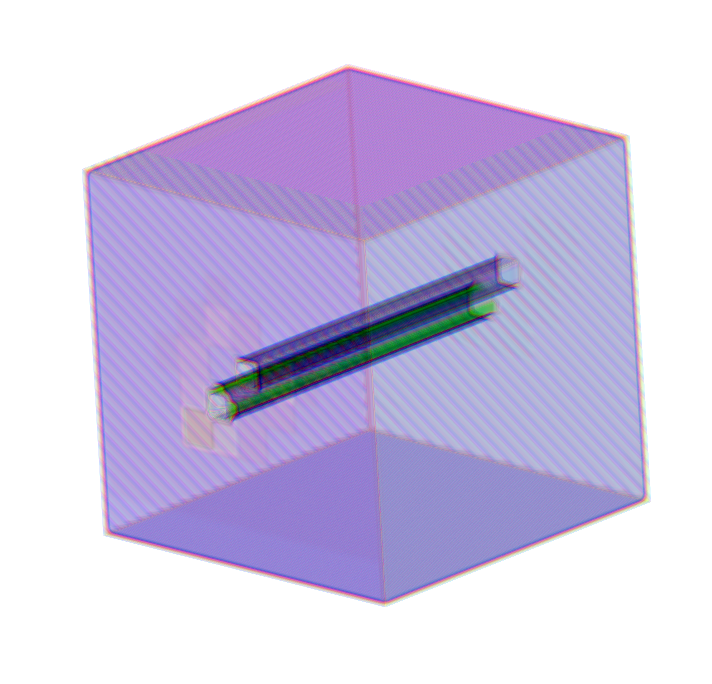}}}

\\
\hline %row 4
\raisebox{-10mm}{\tabincell{l}{L-Block with a hole}}
&
\raisebox{-10mm}{\tabincell{l}{Non-Manufacturable}}
& 
\raisebox{-10mm}{\tabincell{c}{Non-Manufacturable\\ (Feature: Depth/Diameter ratio\\of hole $<$ 5)}}
& 
\raisebox{-10mm}{\tabincell{l}{\includegraphics[width=20mm, height=20mm]{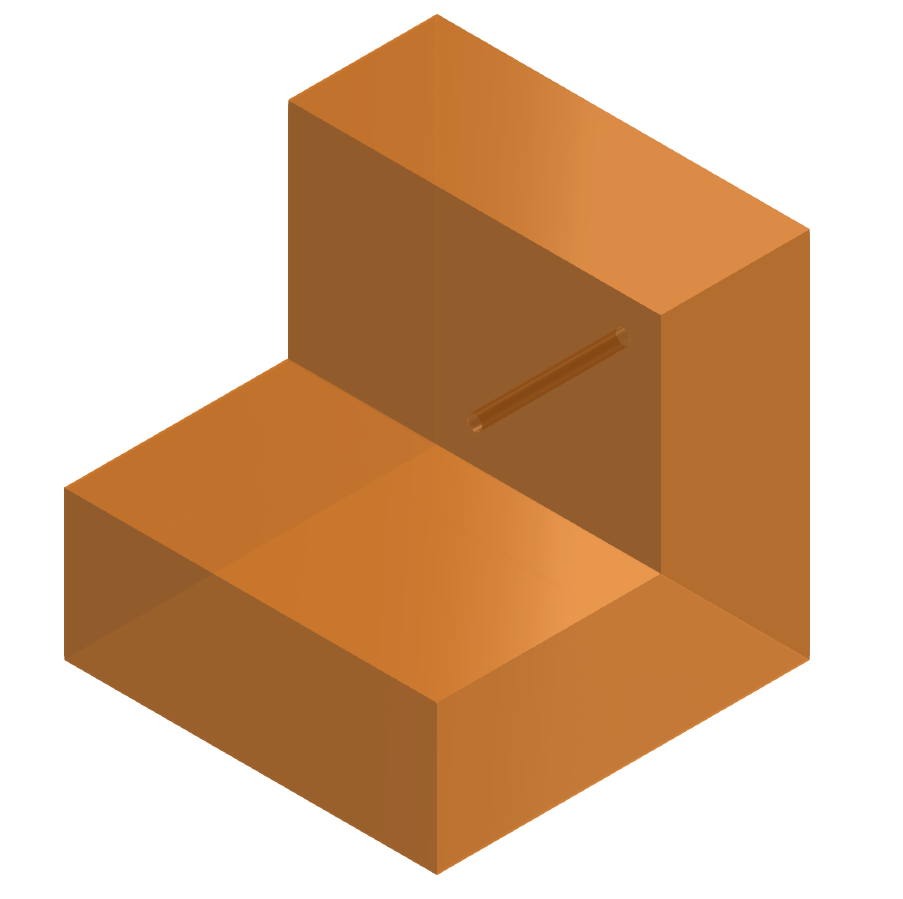}}}
&
\raisebox{-10mm}{\tabincell{l}{\includegraphics[width=20mm, height=20mm]{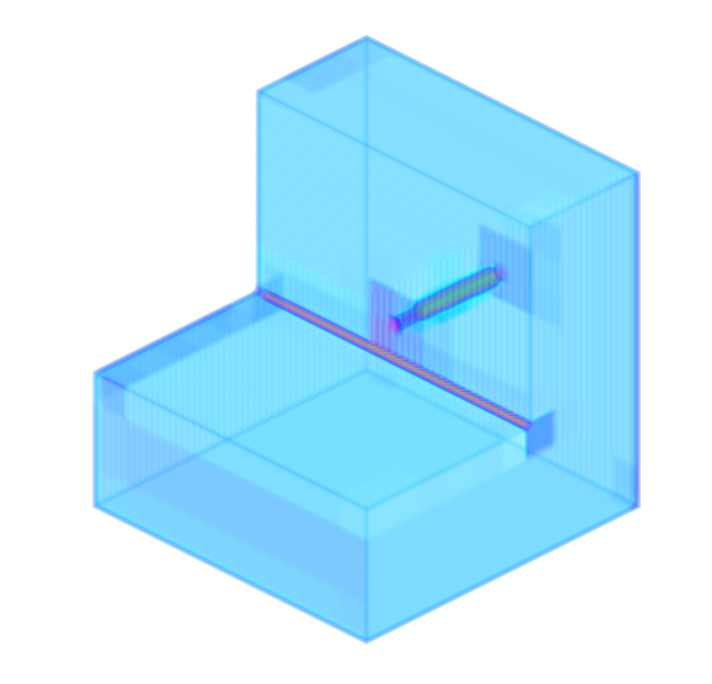}}}

\\
\hline %row 5
\raisebox{-10mm}{\tabincell{l}{L-Block with a hole}}
&
\raisebox{-10mm}{\tabincell{l}{Non-Manufacturable}}
& 
\raisebox{-10mm}{\tabincell{c}{Non-Manufacturable\\ (Feature: Hole close to\\the corner)}}
& 
\raisebox{-10mm}{\tabincell{l}{\includegraphics[width=20mm, height=20mm]{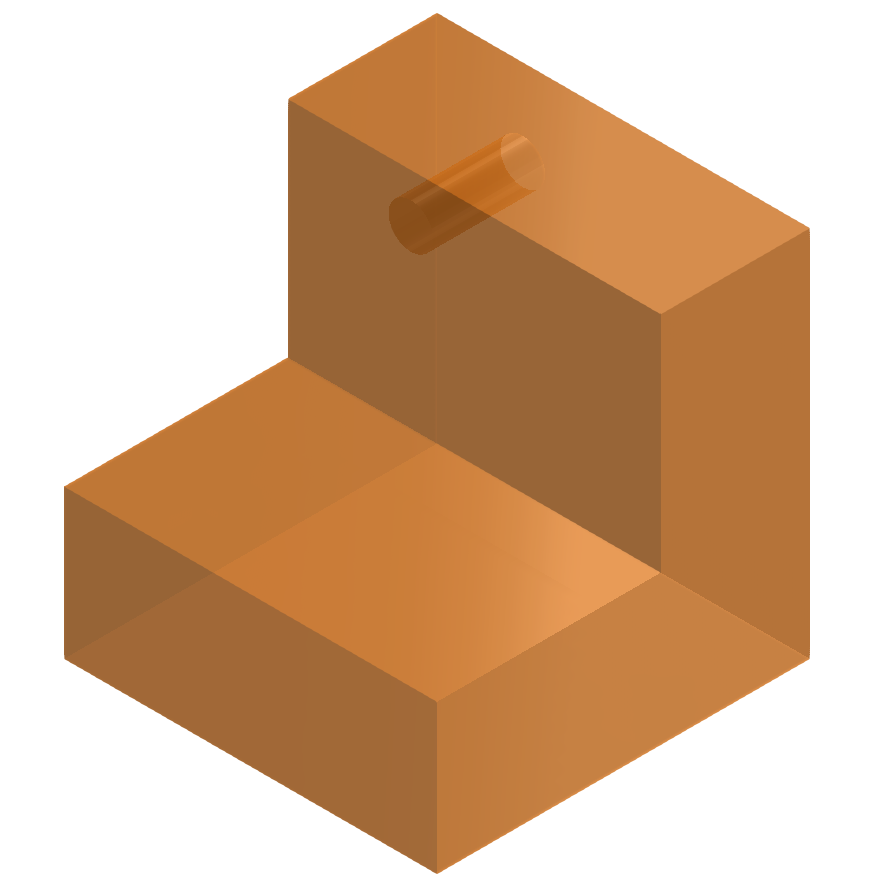}}}
&
\raisebox{-10mm}{\tabincell{l}{\includegraphics[width=20mm, height=20mm]{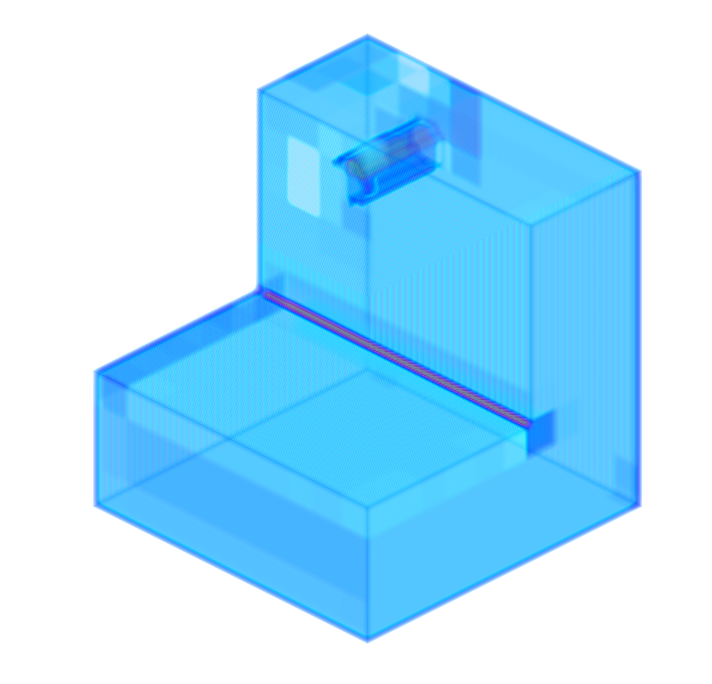}}}

\\
\hline % row 6
\raisebox{-10mm}{\tabincell{l}{Two holes\\Different faces}}
&
\raisebox{-10mm}{\tabincell{l}{Non-Manufacturable}}
& 
\raisebox{-10mm}{\tabincell{c}{Non-Manufacturable\\(Feature: One hole is\\manufacturable\\One hole is\\non-manufacturable)}}
& 
\raisebox{-10mm}{\tabincell{l}{\includegraphics[width=20mm, height=20mm]{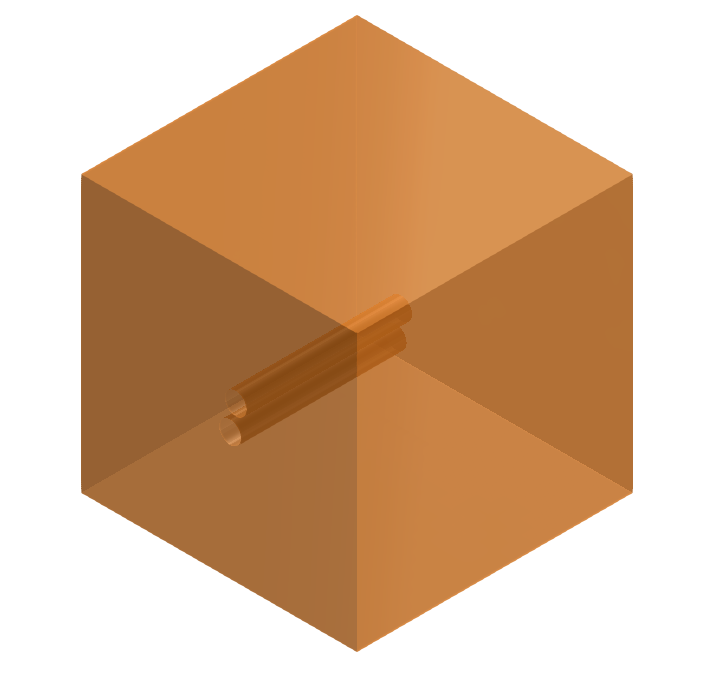}}}
&
\raisebox{-10mm}{\tabincell{l}{\includegraphics[width=20mm, height=20mm]{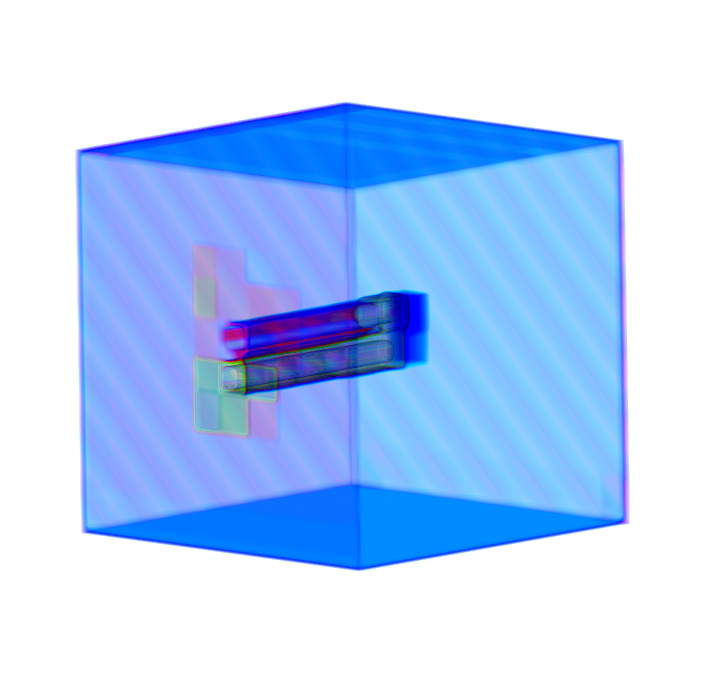}}}

\\
\hline
\raisebox{-10mm}{\tabincell{l}{Two holes\\Same face}}
&
\raisebox{-10mm}{\tabincell{l}{Manufacturable}}
& 
\raisebox{-10mm}{\tabincell{c}{Manufacturable\\(Feature: Depth/Diameter ratio\\ of both holes $>$ 5)}}
& 
\raisebox{-10mm}{\tabincell{l}{\includegraphics[width=20mm, height=20mm]{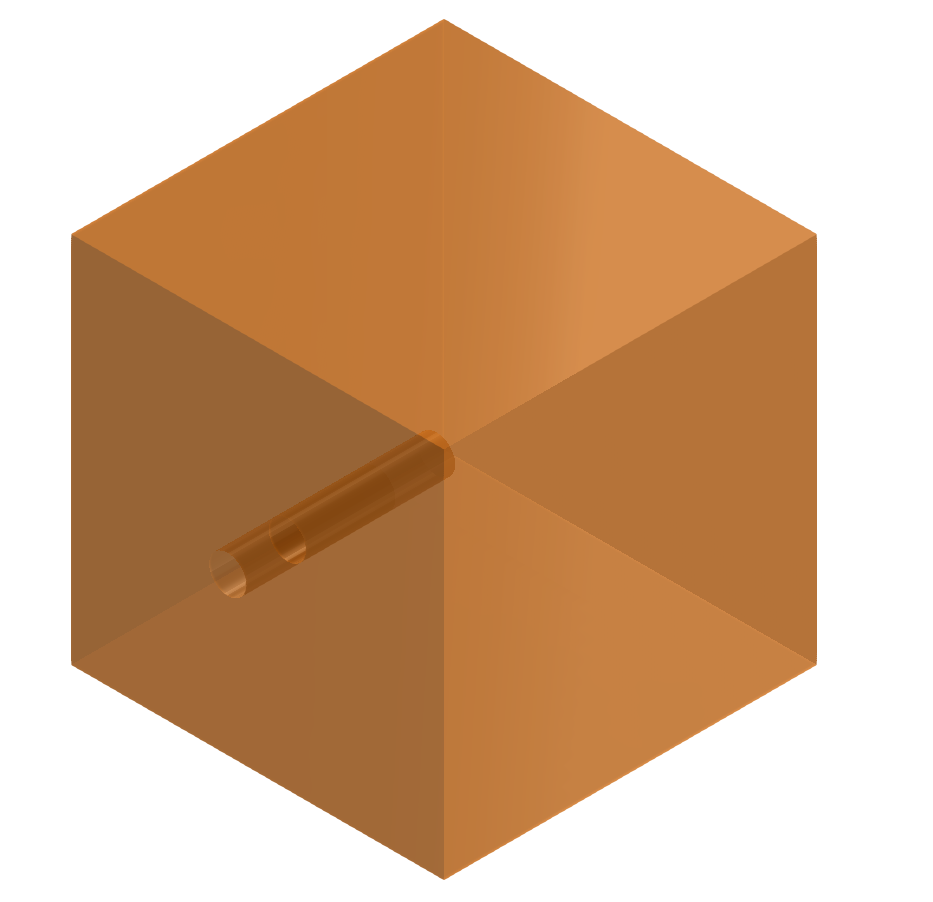}}}
&
\raisebox{-10mm}{\tabincell{l}{\includegraphics[width=20mm, height=20mm]{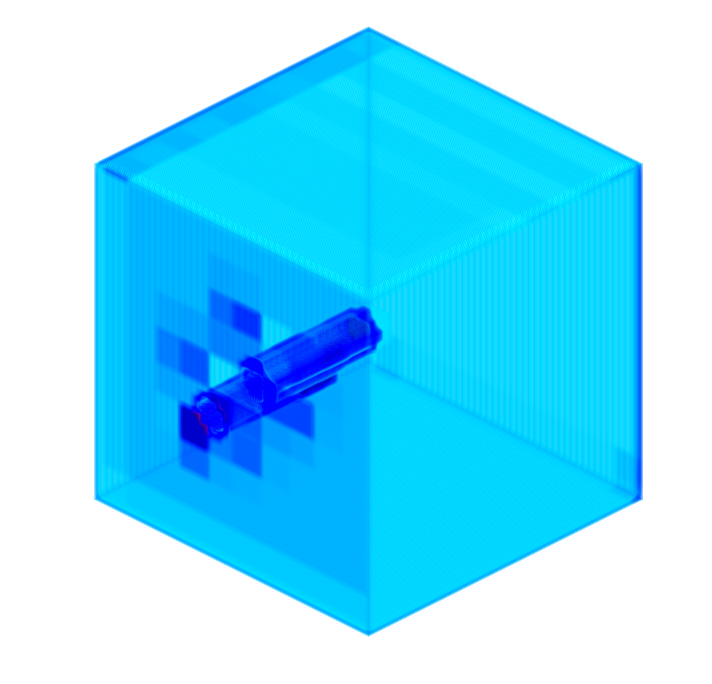}}}

\\
\hline
\raisebox{-10mm}{\tabincell{l}{Cylinder with a hole}}
&
\raisebox{-10mm}{\tabincell{l}{Non-Manufacturable}}
& 
\raisebox{-10mm}{\tabincell{c}{Non-Manufacturable\\(Feature: Hole close to\\the edge)}}
& 
\raisebox{-12mm}{\tabincell{l}{\includegraphics[width=20mm, height=20mm]{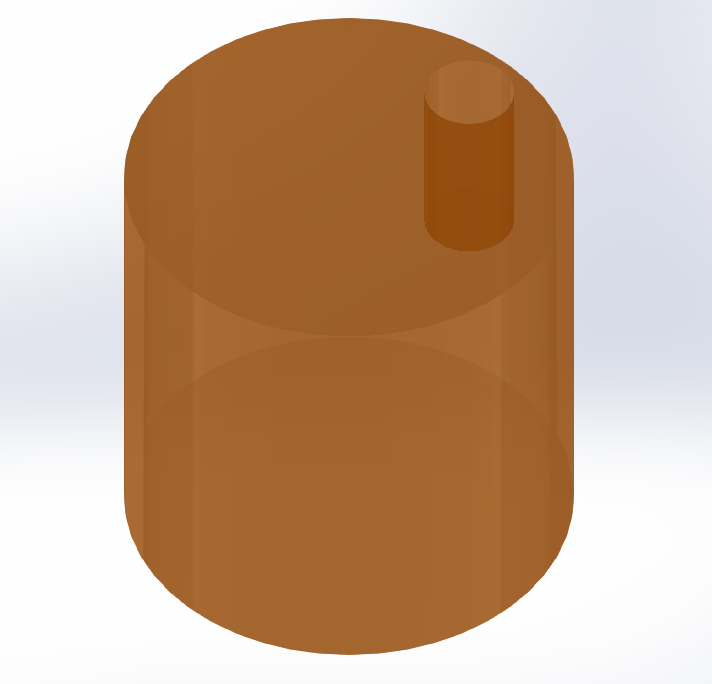}}}
&
\raisebox{-12mm}{\tabincell{l}{\includegraphics[width=23mm, height=23mm]{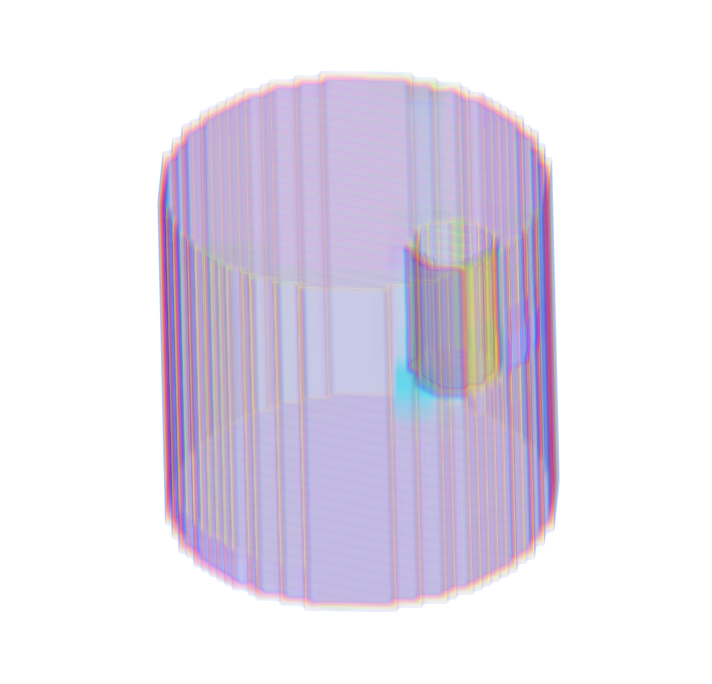}}}

\\
\hline
\end{tabulary}

\caption{Illustrative examples of manufacturability prediction and interpretation using the DLDFM framework.}
\label{Tab:GradCAM}
\end{table*}

\subsection{Visualization of Feature Maps of 3D-CNN}
The ability of the 3D-CNN to learn the local geometric features can be understood by visualizing the feature maps. We provide different sample CAD models to the input of the 3D-CNN and plot the feature maps obtained as output of first layer. The activations are first normalized to \emph{one} and then the activation tensor for each filter is plotted using the \emph{Smooth3} function in MATLAB\textsuperscript{\textregistered}. Figure~\ref{Fig:fm_comp} shows the output obtained from one of the hidden layers of the DLDFM network. It can be seen that the 3D-CNN is able to recognize the primitive information about the geometry; for example, the edges, the faces, the hole, the depth of the hole etc.

\subsection{Visualization of 3D-GradCAM}

Using the trained DLDFM network, it is possible to obtain the localization of the feature activating the decision of the DLDFM as explained in the Section~\ref{Sec:GradCAM}. The 3D-GradCAM renderings for various cases are shown in the Table~\ref{Tab:GradCAM}. We have used 3D-GradCAM to visualize the results of various inputs such as manufacturable holes, non-manufacturable-holes, multiple holes in same face, holes in multiple faces of the cube, L shaped block, and cylindrical block. 3D-GradCAM can localize the features that can cause the part to be non-manufacturable. For example, in Table~\ref{Tab:GradCAM}, the second example from the top shows a CAD model with a hole, which is non-manufacturable because it is too close to one of side faces. This is a difficult example to classify based only on the information of the hole. The fifth example from the top shows a L-shaped block with a hole at its corner which is non-manufacturable due to its proximity to the edges of the block. The 3D-GradCAM rendering correctly identifies the non-manufacturable hole and as a result the DLDFM network also predicts the part to be non-manufacturable. The feedback is helpful to understand which particular feature among various other features in a CAD geometry that accounts for the non-manufacturability and possibly modify the design appropriately. Finally, the last example in Table~\ref{Tab:GradCAM} shows a non-manufacturable hole inside a cylindrical external geometry. This shows that the 3D-CNN can identify local features irrespective of the external shape of the CAD model. This is again useful to employ the manufacturability analysis.

\section{Conclusions and Future work}
In this paper, we demonstrate the feasibility of using 3D-CNNs to identify local features of interest using a voxel-based approach. The 3D-CNN was able to learn local geometric features directly from the voxelized model, without any additional shape information. In addition, augmenting the voxel information with surface normals, helped improve the ability of the 3D-CNN to identify local features. As a result, the 3D-CNN was able to identify the local geometric features irrespective of the external object shape, even in the non-representative test data. Hence, a 3D-CNN can be used effectively to identify local features, which in turn can be used to define a metric that may be used for successful object classification.

We apply our novel local fetaure detection tool to build a deep-learning-based DFM (DLDFM) framework which is a novel application of deep learning for cyber-enabled manufacturing. To the best of our knowledge, this is the first application of deep learning to learn the different DFM rules associated with design for manufacturing. In this paper, our DLDFM framework was able to successfully learn the complex DFM rules for drilling, which include not only the depth-to-diameter ratio of the holes but also their position and type (through hole vs. blind). As a consequence, the DLDFM framework out-performs traditional rule-based DFM tools currently available in CAD systems. The framework can be extended to learn manufacturable features for a variety of manufacturing processes such as milling, turning, etc. We envision training multiple networks for specific manufacturing processes, which can be concurrently used to classify the same design with respect to their manufacturability using different processes. Thus, an interactive decision-support system for DFM can be integrated with current CAD systems, which can provide real-time manufacturability analysis while the component is being designed. This would decrease the design time, leading to significant cost-savings.

\comment{
This paper also serves as a proof-of-concept to demonstrate the feasibility of using 3D-CNNs for manufacturing applications using a voxel-based approach. The DLDFM was able to learn features directly from the voxelized model, without any additional shape information. This shows that the same framework can be extended to learn manufacturable features for a variety of manufacturing processes such as milling, turning, etc. In addition, the same voxelized approach can be used to identify non-manufacturable features for additive manufacturing. We envision training multiple networks for specific manufacturing processes, which can be concurrently used to classify the same design with respect to their manufacturability using different processes. Thus, an interactive decision-support system for DFM can be integrated with current CAD systems, which can provide real-time manufacturability analysis while the component is being designed. This would decrease the design time, leading to significant cost-savings.

\begin{table}[h]
	\newcommand\T{\rule{0pt}{2.6ex}}
	\newcommand\B{\rule[-1.2ex]{0pt}{0pt}}
	\centering
	\tymin=.1in
	\tymax=1.5in 

	\begin{tabulary}{3.2in}{|L|R|R|}
		\hline
          Confusion Matrix & Predicted Manufacturable  &Predicted Non-Manufacturable\\   
          \hline                
          Target Manufacturable\T\B & 2222 & 126 \\
          \hline
          Target Non-Manufacturable\T\B & 153 & 1731 \\
          \hline
	\end{tabulary}
	\caption{Confusion Matrix for the prediction on the test dataset.}
    \label{Tab:conf_dataset3}
\end{table}
}

{\small
\bibliographystyle{IEEEtranN}
\bibliography{CM,ML,CAD}

% Generated by IEEEtranN.bst, version: 1.14 (2015/08/26)
\begin{thebibliography}{25}
\providecommand{\natexlab}[1]{#1}
\providecommand{\url}[1]{#1}
\csname url@samestyle\endcsname
\providecommand{\newblock}{\relax}
\providecommand{\bibinfo}[2]{#2}
\providecommand{\BIBentrySTDinterwordspacing}{\spaceskip=0pt\relax}
\providecommand{\BIBentryALTinterwordstretchfactor}{4}
\providecommand{\BIBentryALTinterwordspacing}{\spaceskip=\fontdimen2\font plus
\BIBentryALTinterwordstretchfactor\fontdimen3\font minus
  \fontdimen4\font\relax}
\providecommand{\BIBforeignlanguage}[2]{{%
\expandafter\ifx\csname l@#1\endcsname\relax
\typeout{** WARNING: IEEEtranN.bst: No hyphenation pattern has been}%
\typeout{** loaded for the language `#1'. Using the pattern for}%
\typeout{** the default language instead.}%
\else
\language=\csname l@#1\endcsname
\fi
#2}}
\providecommand{\BIBdecl}{\relax}
\BIBdecl

\bibitem[Sarkar et~al.(2015)Sarkar, Venugopalan, Reddy, Giering, Ryde, and
  Jaitly]{SVRG14}
S.~Sarkar, V.~Venugopalan, K.~Reddy, M.~Giering, J.~Ryde, and N.~Jaitly,
  ``Occlusion edge detection in rgb-d frames using deep convolutional
  networks,'' \emph{Proceedings of IEEE High Performance Exterme Computing
  Conference, Waltham, MA}, 2015.

\bibitem[Lee et~al.(2009)Lee, Grosse, Ranganath, and Ng]{lee2009convolutional}
H.~Lee, R.~Grosse, R.~Ranganath, and A.~Y. Ng, ``Convolutional deep belief
  networks for scalable unsupervised learning of hierarchical
  representations,'' in \emph{Proceedings of the 26th annual international
  conference on machine learning}.\hskip 1em plus 0.5em minus 0.4em\relax ACM,
  2009, pp. 609--616.

\bibitem[Lore et~al.(2016)Lore, Akintayo, and Sarkar]{lore2015llnet}
K.~G. Lore, A.~Akintayo, and S.~Sarkar, ``Llnet: A deep autoencoder approach to
  natural low-light image enhancement,'' \emph{Pattern Recognition}, 2016.

\bibitem[Larochelle and Bengio(2008)]{HY08}
H.~Larochelle and Y.~Bengio, ``Classification using discriminative restricted
  boltzmann machines,'' in \emph{Proceedings of the 25\textsuperscript{th}
  international conference on Machine learning}.\hskip 1em plus 0.5em minus
  0.4em\relax ACM, 2008, pp. 536--543.

\bibitem[Maturana and Scherer(2015{\natexlab{a}})]{maturana20153d}
D.~Maturana and S.~Scherer, ``{3D} convolutional neural networks for landing
  zone detection from lidar,'' in \emph{2015 IEEE International Conference on
  Robotics and Automation (ICRA)}.\hskip 1em plus 0.5em minus 0.4em\relax IEEE,
  2015, pp. 3471--3478.

\bibitem[Maturana and Scherer(2015{\natexlab{b}})]{maturana2015voxnet}
------, ``Voxnet: A 3d convolutional neural network for real-time object
  recognition,'' in \emph{Intelligent Robots and Systems (IROS), 2015 IEEE/RSJ
  International Conference on}.\hskip 1em plus 0.5em minus 0.4em\relax IEEE,
  2015, pp. 922--928.

\bibitem[Riegler et~al.(2016)Riegler, Ulusoys, and Geiger]{riegler2016octnet}
G.~Riegler, A.~O. Ulusoys, and A.~Geiger, ``Octnet: Learning deep 3d
  representations at high resolutions,'' \emph{arXiv preprint
  arXiv:1611.05009}, 2016.

\bibitem[Huang and You(2016)]{huang2016point}
J.~Huang and S.~You, ``Point cloud labeling using 3d convolutional neural
  network,'' in \emph{Proc. of the International Conf. on Pattern Recognition
  (ICPR)}, vol.~2, 2016.

\bibitem[Wang and Siddiqi(2016)]{wang2016differential}
C.~Wang and K.~Siddiqi, ``Differential geometry boosts convolutional neural
  networks for object detection,'' in \emph{Proceedings of the IEEE Conference
  on Computer Vision and Pattern Recognition Workshops}, 2016, pp. 51--58.

\bibitem[Shukor and Axinte(2009)]{shukor2009manufacturability}
S.~A. Shukor and D.~Axinte, ``Manufacturability analysis system: issues and
  future trends,'' \emph{International Journal of Production Research},
  vol.~47, no.~5, pp. 1369--1390, 2009.

\bibitem[Krishnamurthy et~al.(2009)Krishnamurthy, Khardekar, and
  McMains]{Krishnamurthy-2009-CAD}
A.~Krishnamurthy, R.~Khardekar, and S.~McMains, ``Optimized {GPU} evaluation of
  arbitrary degree nurbs curves and surfaces,'' \emph{Computer-Aided Design},
  vol.~41, no.~12, pp. 971--980, 2009.

\bibitem[Riegler et~al.(2017)Riegler, Ulusoy, Bischof, and
  Geiger]{riegler2017octnetfusion}
G.~Riegler, A.~O. Ulusoy, H.~Bischof, and A.~Geiger, ``Octnetfusion: Learning
  depth fusion from data,'' \emph{arXiv preprint arXiv:1704.01047}, 2017.

\bibitem[Tatarchenko et~al.(2017)Tatarchenko, Dosovitskiy, and
  Brox]{tatarchenko2017octree}
M.~Tatarchenko, A.~Dosovitskiy, and T.~Brox, ``Octree generating networks:
  Efficient convolutional architectures for high-resolution 3d outputs,''
  \emph{arXiv preprint arXiv:1703.09438}, 2017.

\bibitem[Ji et~al.(2013)Ji, Xu, Yang, and Yu]{ji20133d}
S.~Ji, W.~Xu, M.~Yang, and K.~Yu, ``3d convolutional neural networks for human
  action recognition,'' \emph{IEEE transactions on pattern analysis and machine
  intelligence}, vol.~35, no.~1, pp. 221--231, 2013.

\bibitem[Tran et~al.(2015)Tran, Bourdev, Fergus, Torresani, and
  Paluri]{tran2015learning}
D.~Tran, L.~Bourdev, R.~Fergus, L.~Torresani, and M.~Paluri, ``Learning
  spatiotemporal features with 3d convolutional networks,'' in \emph{2015 IEEE
  International Conference on Computer Vision (ICCV)}.\hskip 1em plus 0.5em
  minus 0.4em\relax IEEE, 2015, pp. 4489--4497.

\bibitem[Payan and Montana(2015)]{payan2015predicting}
A.~Payan and G.~Montana, ``Predicting alzheimer's disease: a neuroimaging study
  with 3d convolutional neural networks,'' \emph{arXiv preprint
  arXiv:1502.02506}, 2015.

\bibitem[Lee et~al.(2015)Lee, Zlateski, Vishwanathan, and
  Seung]{lee2015recursive}
K.~Lee, A.~Zlateski, A.~Vishwanathan, and H.~S. Seung, ``Recursive training of
  2d-3d convolutional networks for neuronal boundary detection,'' \emph{arXiv
  preprint arXiv:1508.04843}, 2015.

\bibitem[Hinton and Salakhutdinov(2006)]{hinton2006reducing}
G.~E. Hinton and R.~R. Salakhutdinov, ``Reducing the dimensionality of data
  with neural networks,'' \emph{Science}, vol. 313, no. 5786, pp. 504--507,
  2006.

\bibitem[Zeiler(2012)]{zeiler2012adadelta}
M.~D. Zeiler, ``Adadelta: an adaptive learning rate method,'' \emph{arXiv
  preprint arXiv:1212.5701}, 2012.

\bibitem[Selvaraju et~al.(2016)Selvaraju, Das, Vedantam, Cogswell, Parikh, and
  Batra]{selvaraju2016grad}
R.~R. Selvaraju, A.~Das, R.~Vedantam, M.~Cogswell, D.~Parikh, and D.~Batra,
  ``Grad-cam: Why did you say that? visual explanations from deep networks via
  gradient-based localization,'' \emph{arXiv preprint arXiv:1610.02391}, 2016.

\bibitem[Boothroyd et~al.(2002)Boothroyd, Dewhurst, and
  Knight]{BoothroydDewhurst}
G.~Boothroyd, P.~Dewhurst, and W.~Knight, \emph{{Product Design for Manufacture
  and Assembly}}.\hskip 1em plus 0.5em minus 0.4em\relax M. Dekker, 2002.

\bibitem[Bralla(1999)]{bralla1999design}
J.~G. Bralla, \emph{Design for manufacturability handbook}.\hskip 1em plus
  0.5em minus 0.4em\relax McGraw-Hill,, 1999.

\bibitem[{Spatial Corporation}(2009)]{ACIS10}
{Spatial Corporation}, \emph{{ACIS} Geometric Modeler: User Guide}, 2009,
  version 20.0.

\bibitem[Chollet(2015)]{chollet2015keras}
F.~Chollet, ``Keras,'' \url{https://github.com/fchollet/keras}, 2015.

\bibitem[Abadi et~al.(2015)Abadi, Agarwal, Barham, Brevdo, Chen, Citro,
  Corrado, Davis, Dean, Devin, Ghemawat, Goodfellow, Harp, Irving, Isard, Jia,
  Jozefowicz, Kaiser, Kudlur, Levenberg, Man\'{e}, Monga, Moore, Murray, Olah,
  Schuster, Shlens, Steiner, Sutskever, Talwar, Tucker, Vanhoucke, Vasudevan,
  Vi\'{e}gas, Vinyals, Warden, Wattenberg, Wicke, Yu, and
  Zheng]{tensorflow2015-whitepaper}
\BIBentryALTinterwordspacing
M.~Abadi, A.~Agarwal, P.~Barham, E.~Brevdo, Z.~Chen, C.~Citro, G.~S. Corrado,
  A.~Davis, J.~Dean, M.~Devin, S.~Ghemawat, I.~Goodfellow, A.~Harp, G.~Irving,
  M.~Isard, Y.~Jia, R.~Jozefowicz, L.~Kaiser, M.~Kudlur, J.~Levenberg,
  D.~Man\'{e}, R.~Monga, S.~Moore, D.~Murray, C.~Olah, M.~Schuster, J.~Shlens,
  B.~Steiner, I.~Sutskever, K.~Talwar, P.~Tucker, V.~Vanhoucke, V.~Vasudevan,
  F.~Vi\'{e}gas, O.~Vinyals, P.~Warden, M.~Wattenberg, M.~Wicke, Y.~Yu, and
  X.~Zheng, ``{TensorFlow}: Large-scale machine learning on heterogeneous
  systems,'' 2015, software available from tensorflow.org. [Online]. Available:
  \url{http://tensorflow.org/}
\BIBentrySTDinterwordspacing

\end{thebibliography}
}

\end{document}